\pdfoutput=1
\documentclass[sn-chicago]{sn-jnl}

\usepackage{graphicx}%
\usepackage{multirow}%
\usepackage{amsmath,amssymb,amsfonts}%
\usepackage{amsthm}%
\usepackage{mathrsfs}%
\usepackage[title]{appendix}%
\usepackage{xcolor}%
\usepackage{textcomp}%
\usepackage{manyfoot}%
\usepackage{booktabs}%
\usepackage{algorithm}%
\usepackage{algorithmicx}%
\usepackage{algpseudocode}%
\usepackage{listings}%
\usepackage{multicol}
\usepackage{subcaption}
\usepackage{listings}

\begin{document}

\title[Italian Embeddings for Concepts Description]{Domain Embeddings for Generating Complex Descriptions of Concepts in Italian Language}


\author*{\fnm{Alessandro} \sur{Maisto}}\email{amaisto@unisa.it}

\affil{\orgdiv{Department of Politics and Communication Science}, \orgname{University of Salerno}, \orgaddress{\street{via Giovanni Paolo II, 132}, \city{Fisciano (SA)}, \postcode{84084}, \state{Italia}, \country{Country}}}


\abstract{In this work, we propose a Distributional Semantic resource enriched with linguistic and lexical information extracted from electronic dictionaries, designed to address the challenge of bridging the gap between the continuous semantic values represented by distributional vectors and the discrete descriptions offered by general semantics theory. Recently, many researchers have concentrated on the nexus between embeddings and a comprehensive theory of semantics and meaning. This often involves decoding the representation of word meanings in Distributional Models into a set of discrete, manually constructed properties such as semantic primitives or features, using neural decoding techniques. Our approach introduces an alternative strategy grounded in linguistic data. We have developed a collection of domain-specific co-occurrence matrices, derived from two sources: a classification of Italian nouns categorized into 4 semantic traits and 20 concrete noun sub-categories, and a list of Italian verbs classified according to their semantic classes. In these matrices, the co-occurrence values for each word are calculated exclusively with a defined set of words pertinent to a particular lexical domain. The resource comprises 21 domain-specific matrices, one comprehensive matrix, and a Graphical User Interface. Our model facilitates the generation of reasoned semantic descriptions of concepts by selecting matrices directly associated with concrete conceptual knowledge, such as a matrix based on location nouns and the concept of animal habitats. We assessed the utility of the resource through two experiments, achieving promising outcomes in both: the automatic classification of animal nouns and the extraction of animal features.}




\maketitle

\section{Introduction}\label{sec1}

In the last two decades, Distributional Semantics (DS) have reached wild popularity among computational linguists. The popularity of the DS models, also called ’Word Embeddings’, can be due mainly to two factors: the general idea behind DS models is fairly straightforward, as well as it is easy to read and produce results with them; Distributional Semantics’ main idea, words with similar contexts have similar meanings, has a strong anchor on theoretical linguistics, in particular with Harris’ Distributional Hypothesis \citep{harris1954distributional}.\\
Distributional Semantics models represent a concretisation of Harris’ theory. They work by building large co-occurrence matrices and comparing word vectors to generate semantic similarity values among words automatically.

Nevertheless, many authors have recently focused on connecting embeddings and a general theory of semantics and meaning. The classic semantic features representation of a concept consists of a set of discrete, finite, manually generated properties which can be formalised, for example, as follows:

\begin{itemize}
    \item[] \textbf{Tiger}
    \begin{itemize}
        \item is\_a\_feline
        \item has\_tail
        \item has\_fur
        \item has\_stripes
        \item is\_yellow
        \item is\_black
    \end{itemize}
\end{itemize}

Contrariwise, Distributional Semantics represents concepts as an extensive list of cooccurrence values stored in vectors, which can be considered continuous due to its vast number of dimensions \citep{chersoni2021decoding}. As \cite{chersoni2021decoding} pointed out, the problem is the \textit{lack of interpretability} of embeddings, which \textit{provide a} holistic \textit{representation of meaning}. In such a way, the semantic content of a word can only be described in association with other elements in the semantic space. For example, the concept of \textit{tiger} is defined by the similarity relations between the word \textit{tiger} and all the other words. The similarity values are high with \textit{leopard}, \textit{lion} or \textit{cat} while very low with many other words.\\
The coordinates of a word in that semantic space are completely arbitrary and devoid of an intrinsic semantic value \citep{sahlgren2008distributional}. Their meanings are \textit{caught in a circle of similar words} \citep{murphy2004big}.\\
\cite{boleda2020distributional} propose to decode the continuous representation of words’ meaning of Distributional Models in a set of discrete and manually constructed properties like semantic primitives or features. \citeauthor{boleda2020distributional} underlines that Distributional representations can be considered attractive as a model of word meaning for three main reasons: they are learned directly from natural language; their representations have high multidimensionality that can encode minimal nuances of meaning; differences between words are expressed in continuous values.

In this work, we propose a linguistic resource for the Italian language that represents an alternative strategy for dealing with the issue of the interpretability of distributional semantic spaces. Instead of trying to establish a connection between vectors’ dimensions and conceptual knowledge, in our approach, we build a set of different, domain-restricted co-occurrence matrices. We do not attempt to interpret the contents of a distributional matrix; instead, we seek to represent the meaning of a word in various matrices that are semantically constrained to a particular semantic field. We train the model over a generic corpus, only looking for word co-occurrences within a given semantic domain. In this way, we can generate a semantically motivated description of concepts by only choosing dimensions specifically related to concrete conceptual knowledge. We use a set of tagged electronic dictionaries, including: 

\begin{itemize}
    \item a dictionary of Italian nouns, which includes semantic tags referring to the lexicosyntactic traits identified by \cite{chomsky1965aspects}. This dictionary contains 19 semantic sub-category tags for concrete nouns.
    \item a selection of three semantic macro-classes of Italian Verbs, which collection started from a review of the semantic classes of predicates identified into the mark of the Lexicon-Grammar theory \citep{gross1981bases}.
\end{itemize}

We tested the capacity of the proposed set of matrices to represent specific semantic meanings with two experiments related to the animal domain semantics. First, we created a classification of nouns associated with zoology study. To accomplish this, we assume that what is valid for sentences in the so-called Compositional Distributional Semantics \citep{baroni2014frege} is also true for the meaning of a single word. We define a word as a tensor by combining a set of matrices that reflect the semantic features of a concept.

Second, we carried out an experiment of automatic semantic features generation in which we combined different domain matrices to extract similarity values among generic animal nouns and nouns of prototypical animals.

The paper is structured as follows: section 2 presents the related works; Section 3 will include an accurate description of the proposed resource and its production; Section 4 will describe the experiments and show the results; in Section 5, we will offer some final considerations.

\section{State of the Art}
The problem of establishing a connection between semantic theories and Distributional models dates back to the first Distributional Models. \cite{murphy2004big} pointed out that the growing availability of fast processors and large computer memories had impulse the development of Distributional Semantic analysis. By analysing the results of two of the first models, LSA \citep{landauer1997solution} and HAL \citep{burgess1998simple}, Murphy acknowledged the effectiveness of this approach but also pointed out their limits. They were based on an associationistic approach to the psychological model of word use that cannot replace the conceptual approach. Nevertheless, Murphy also shows a possible way forward for future DS models by turning the numerical dimensions of a word vector into meaningful semantic dimensions. As Murphy wrote, \textit{rather than just providing a global measure of similarity between cat and horse, LSA might allow us to examine the dimensions on which they are similar and dissimilar and thereby specify the properties of cats and horses... At that point, LSA would not necessarily be a different theory of word meaning but could be seen as a way of implementing conceptual knowledge} \cite[p.~430]{murphy2004big}.

Murphy’s idea of decoding the semantic content of word embeddings has been carried on in the last year by different authors. \cite{boleda2015distributional} suggest that distributional semantics representations allow making inferences in a similar way semantic primitives do. They \textit{capture information that is analogous to the information primitives were designed to capture}. Since Distributional Semantics representations are learned on natural data (the context of the word taken in texts naturally produced by humans), they \textit{can serve as the basis for a semantic representation of words and phrases that serve many of the purposes semantic primitives were designed for, without running into many of their philosophical, empirical, and practical problems}. But distributional approaches also face critical problems: no clear understanding of distributional inference exists. With this goal, \citep{mikolov2013linguistic} used vector combination: if we subtract the vector of man from the vector of \textit{king} and add the vector of \textit{woman} to the difference, we obtain a new vector that is very close to the vector of the word \textit{queen}. In addition, primitives inherit the notion of compositionality from logic, while distributional approaches to compositionality are not optimised. Another significant problem is that distributional representation doesn’t have a relation with the physical world  \citep{fagarasan2015distributional}. In DS models, words' meanings are defined primarily in connection to other words, even though the semantics of words are also formed from our interactions with the outside world.

\citeauthor{boleda2015distributional} propose using a hybrid approach that combines human-generated features with embeddings. Nevertheless, many authors point out the problem of the lack of manually generated semantic feature norms  \citep{fagarasan2015distributional,derby2019feature2vec,chersoni2021decoding}. 

\cite{fagarasan2015distributional} and \cite{derby2019feature2vec} tried to establish a map between the feature domain and the embedding space. In particular, \citeauthor{fagarasan2015distributional} focused on predicting concept features by mapping a distributional space to a cognitive-based semantic space. On the contrary, \citeauthor{derby2019feature2vec} mapped from the feature domain to embedding space. Both experiments used \cite{mcrae2005semantic} data set as the cognitive set of semantic features. 

In \cite{utsumi2018neurobiologically}, the authors attempt to discover the internal knowledge of embedded word representations for cognitive modelling. They proposed a computational experiment using the featural conceptual representation presented in \cite{binder2016toward}. They ground the assertion that DSMs \textit{reflect the representational structure of semantic knowledge in the brain} on numerous studies on brain imaging that used distributional word vectors to predict brain activity in lexical processing \citep{mitchell2008vector}.\\
\cite{anderson2017visually,pereira2018toward} also tried to simulate human conceptual representation collected in  \citeauthor{binder2016toward} with the information encoded in a word vector. Their results demonstrate that abstract knowledge is encoded in word vectors better than spatiotemporal and sensorimotor information. Among abstract information, cognitive, social, causal and emotional information has been simulated with better results.

The approach of \cite{chersoni2021decoding} is similar to that of \citeauthor{utsumi2018neurobiologically}. Nevertheless, \citeauthor{chersoni2021decoding} apply a neural decoding methodology to contextualised embeddings. Generally, neuro-semantic decoding is used to  \textit{infer the content of brain activities associated with certain words or phrases} as proposed by \citeauthor{mitchell2008vector}. A common approach consists of mapping fMRI signal vectors onto semantic dimension vectors. In \cite{chersoni2017structure}, neuro-semantic decoding aims \textit{to decode the content of word embeddings themselves}.

Other influential studies that followed the method proposed by \citeauthor{mitchell2008vector} tested the capacity of word embeddings to predict brain activity focused on count-based dependency models with linear regression \citep{murphy2012selecting} or skip-gram and other count-based models using regression and similarity-based encoding and decoding methods \citep{anderson2016representational}. Those methods use human subjects to generate brain activation patterns in specific semantic tasks (they were asked to think about features of a particular concept) with fMRI and then compare those patterns to the distributional vector of the target word.

Differently from the model previously presented, in our approach, the interpretation of word embeddings is provided \textit{a priori} since we forced the semantic general content of the dimension of the matrix by setting up a controlled domain dictionary in the steps previous to its training.

Our resource comprises a set of freely downloadable matrices and an easy user interface.

The idea of creating a general framework from which downloading a set of matrices, each of which specialised in solving a specific problem, was introduced by \cite{baroni2010distributional} with Distributional Memory. The authors released the resource\footnote{https://marcobaroni.org/dm/} as a framework that extracts distributional information from the corpus as a set of weighted word-link-word tuples organised into a third-order tensor. The tensor is then used to build various matrices, and their rows and columns represent natural spaces for dealing with various semantic problems. Our approach strongly differs from Distributional Memory because we directly released the different matrices not yet combined in a tensor.

There are many other open-source Distributional Semantic resources on the web, but, in general, they are challenging to use for a no-specialized user. Many of these require a minimum knowledge of a programming language. Nevertheless, there are many resources which are used extensively.

Among the most recent resources, the most known is BERT \citep{devlin2018}, whose pre-trained models have been released by Google Research Group\footnote{https://github.com/google-research/bert} \citep{turc2019well}. BERT is a contextualised model that works by learning the vectors as a function of the internal states of a Transformer. As a contextualised model, it can build different vectors for each meaning of a single noun by differentiating the different contexts in which the word appears. \\
Other influential resources are GloVe \citep{pennington2014glove} and Word2Vec \citep{mikolov2013efficient}, both released in the Gensim python library\footnote{https://radimrehurek.com/gensim/index.html}. Glove builds the matrix by calculating the co-occurrence values as the difference of the conditional probability that both words appear with other words. Contrariwise, Word2Vec utilises a neural network model to discover word associations from text corpora.\\
\cite{jurgens2010s} proposed an interesting collection of models. They developed a Java package for Distributional Semantics called S-Space\footnote{https://github.com/fozziethebeat/S-Space/wiki}, which includes many models such as HAL \citep{burgess1998simple}, LSA \citep{landauer1997solution}, Random Indexing \citep{sahlgren2008permutations}, Dependency Vectors \citep{pado2007dependency} among others. Furthermore, the package provides many algorithms which can be included in new models, such as matrix transformation algorithms, semantic similarity metrics, corpus readers, etc...

ALaCarte embeddings \citep{khodak2018carte} is another interesting project. This resource\footnote{https://github.com/NLPrinceton/ALaCarte} represents an alternative to Word2Vec for rare or unusual words. It can also build a semantic representation of a word if the corpus contains only a few examples of that word.

In the next Sessions, we will illustrate the resource, providing two experiments demonstrating its use and potentialities.

\section{Building the set of matrices}
\label{method}
While many authors are working on decrypting the ’black boxes’ represented by word embeddings to find a connection between features and groups of vectors’ dimensions, our work relies on an inverted principle: we aim to create a series of controlled matrices, each featuring a fixed number of specific, domain-related dimensions. These matrices are intended to serve as specialized descriptions of individual features of a given word.

Our work begins by creating a collection of small matrices from a set of tagged dictionaries that can describe concrete/abstract features of a wide range of concepts. As we said in section \ref{sec1}, we use two primary dictionary resources to extract the different sets of words which represent the dimensions of our matrices. First, we use an extensive dictionary of Italian nouns divided into 20 semantic classes and organised into semantic Traits. The second resource is a list of Italian predicates divided into syntactic and semantic classes.

\subsection{The Lexicon-Grammar resources}
The Lexicon-Grammar (LG) is a theoretical-methodological approach proposed by Maurice Gross \citep{gross1968grammaire,gross1975methodes} between the 60s and 70s. 

Gross starts his work on the French lexicon into the Generative theoretical framework. He searches for lexical constraints on syntactic generalised transformations as required by the Generative program \citep{chomsky1965aspects}. Chomsky affirmed that "a full description of the syntax of a language implies not only the identification of general syntactic rules but also, and equally importantly, a detailed specification of which word requires, accepts or forbids the application of which syntactic rule" \citep{gardent2005maurice}.

Gross adopted a structuralist approach, drawing inspiration from Harris’ concept of transformation, whose concretisation into a distributional and transformational classification of verbal predicates \citep{gross1975methodes} become one of the LG basic concepts. Gross "contributed to the revival of formal linguistics in the 1960s, and he created and implemented an efficient methodology for descriptive lexicology." \citep{laporte2005memoriam}. Gross was also a pioneer of linguistic-based NLP, producing a significant amount of electronic linguistic data.

LG methodology is based on three basic statements: the syntax is indissoluble from the lexicon; simple sentences are the minimum syntactic and semantic objects of analysis; the LG formalisation and methodology must be suitable to any language \citep{elia2011concept}.

The first axiom was based on the data Gross collected in his early studies. He systematically analysed thousands of French verbs to construct a Generative Grammar of French. However, when he realised that the exception encoded into the lexicon far exceeded the syntactic rules, he realised that it was impossible to propose a general theory of syntax without considering the contribution of the lexical entries.

The second axiom states that simple sentences are the minimum object of analysis. The lexicon often generates ambiguities that could be solved at the simple sentence level. The verb ’to miss’ can assume different meanings if accompanied by certain words or others. The verb 'to miss' can assume different meanings if accompanied by certain words or by others: in Sentence 1, the verb has a concrete meaning, while in 2, it assumes a psychological connotation.

\begin{enumerate}
    \item The hunter missed his prey
    \item Mary missed her sons
\end{enumerate}

In addition, reviving the concept of kernel sentences drawn from Harris, Gross considered complex sentences as a result of concatenation and reduction of simple sentences, which must represent the first level of investigation.

The third statement is a direct consequence of the LG methodology, which is thought to be a scientific, empirical, rigorous and systematic study of a language's lexicon and syntax \citep{laporte2005memoriam}. Since the lexicon encoded a vast number of exceptions, which in many cases are restricted only to a single language, the study must be systematic. We cannot assume that a whole class of verbs generally accepts a transformation if we have not tested this transformation with all the possible lexical entries which can appear in the sentence. To do this, Gross, inspired by Harris, employed two kinds of analysis: a Distributional analysis, which regards the lexicon and the co-occurrence of the lexical entries into the sentences; a transformational analysis, which regards the applicability of specific transformation to classes and combination of classes of words.

Starting from sentences 1 and 2, the LG methodology analyses how the elements select their distribution: the verb in Sentence 2, for example, must select an animated being as subject (\textit{[Max + the boy + the dog + *the rifle] missed his sons}). In contrast, sentence 1 also selects an instrument or an event (\textit{[Max + the hunter + the missile + the shot] missed its/his target}). The two verbs also differ in how they select the object: in sentence 1, the object is restricted to concrete nouns, but in sentence 2, it is unrestricted and could also accept an \textit{-ing} form (\textit{Mary missed [her sons + her old life + the city + hearing his voice]}).\\
The transformational analysis is required to examine all potential syntactic transformations a sentence can accept, also in relation to the co-occurrence lexicon, i.e. the acceptability of passive transformation of ’to miss’ with specific classes of subjects or objects (\textit{hearing his voice + her sons was missed by Mary}).

The systematic study of the lexicon generated a vast classification of lexical entries in various languages. The starting point, at a syntactic level, is represented by a classification of predicates of French. The Lexicon-Grammar methodology was born as a systematic description of the syntactic properties of the operators of French (verbs, predicative nouns and adverbs). 

Gross organised the lexicon in tables, each containing the syntactic descriptions of a specific syntactic category. Tables group together all the lexical items entering a specific syntactic construction, and, for each item in the table, there is a set of columns which ”specify the syntactic properties of that item either by adding information about its arguments or by identifying a number of transformations the basic subcategorisation frame associated with the table can undergo” \citep{gardent2005maurice}.

\begin{figure}[H]
    \centering
    \includegraphics[width=6cm]{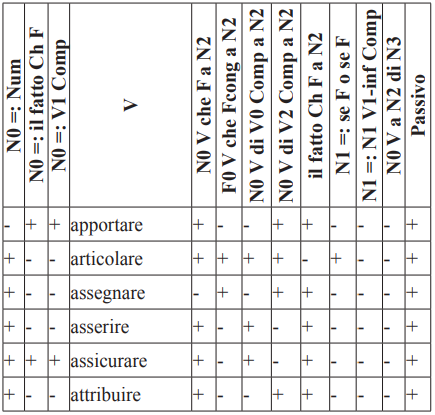}
    \caption{An example of LG Table \citep{elia2010lexis}}
    \label{tables}
\end{figure}

As illustrated by the table in Fig. \ref{tables}, each row of the table corresponds to a lexical entry. Columns contain distributional and transformational features associated or not with the lexical item. Tables are binaries because their items can accept (”+”) or not (”-”) the application of each syntactic property.

Tables are organised on the base of syntactic similarities among verbs. Although no verb has precisely the same properties, the tables are composed of predicates with the same syntactic structure and similarity in specific syntactic behaviour.

\cite{elia1984verbe}was the first to create LG tables for the Italian language, followed by others \citep{elia1981lessico,d1989elaborazione,vietri2004lessico}. Italian LG studies produced a large number of electronic resources which includes the electronic formalisation of the Italian LG Tables of verbs\footnote{\url{http://dsc.unisa.it/composti/tavole/combo/tavole.asp}}, of Frozen Sentences \citep{vietri1990some}\footnote{\url{http://dsc.unisa.it/composti/tavole/sv/table.html}}, of verbs with causal alternation \citep{vietri2017usi}\footnote{\url{http://dsc.unisa.it/tavole/combo/tavole.asp}}, of the manner of motion verbs \citep{vietri2019verbi}\footnote{\url{http://dsc.unisa.it/tavole/sv/vmm/tavole.asp}}, electronic dictionaries and Finite-State Grammars\citep{d2004lexicon,vietri2014italian,de1995dizionario}.

LG produced resources for other parts of speech, such as adjectives, adverbs and nouns, analysing the internal structure of compounds and their combination with the co-occurrent elements. In fact, an extensive distributional analysis needs a complete classification of all the lexical Parts of Speech in groups determined by their semantic similarity. For example, in the analysis of sentences 1 and 2, the LG methodology will classify the different subjects on their semantic class membership (human nouns, instruments, etc.).

\subsection{Dictionaries of Nouns}
\label{conc}
\begin{figure}[H]
    \centering
    \includegraphics[width=8cm]{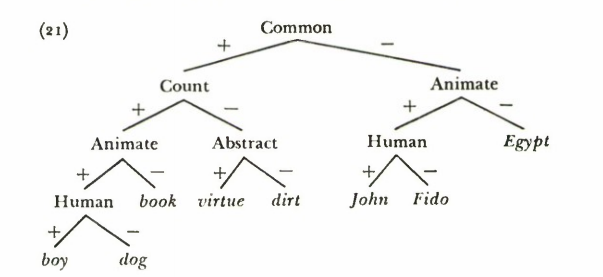}
    \caption{Lexicon-Semantic Traits of Nouns \cite[p.~83]{chomsky1965aspects}}
    \label{fig1}
\end{figure}

The Dictionary of Nouns is part of a more general Italian Dictionary of Simple Words \citep{vietri2014italian} originally built for the software NooJ \citep{silberztein2016formalizing} and then converted into JSON format for lemmatisation purposes \citep{maisto2021building}. From the original dictionary, which has over 129.000 lemmas and more than 1 million flexed forms, we extracted about 21.500 concrete nouns, 3300 proper nouns, 1100 toponyms, 12.000 human nouns and 2.000 animal nouns. Those dictionaries reflect the semantic traits identified by \cite{chomsky1965aspects} (fig. \ref{fig1}). 

Starting at the bottom-left of the tree, we dispose of the resources listed below:

\begin{itemize}
    \item HUMAN+: the list of common Human nouns (\textit{Num} tag);
    \item HUMAN-: the list of Animals Nouns (\textit{Anl} tag);
    \item ANIMATE-: the list of Concrete Nouns (see table \ref{tab:1});
    \item COUNT-ABSTRACT-: the list of Concrete Nouns (\textit{CONC} tag);
    \item COMMON-ANIMATE+: the list of proper nouns (\textit{Npr} tag);
    \item COMMON-ANIMATE-: the list of Toponyms (\textit{TOPONIMO} tag).
\end{itemize}

Non-animate common nouns compose the Concrete Nouns dictionary, an extensive resource sub-categorized into the 20 semantic classes shown in Table \ref{tab:1}. 

The dictionaries allow multiple tagging because some nouns can be associated with different classes. An example of dictionary entries is the following:

\begin{lstlisting}
 [
  "scuola", 
  {
   "num": "s", 
   "gen": "f", 
   "POS": "N", 
   "lemma": "scuola", 
   "token": "scuola", 
   "SEM": ["Num","Nedi"],
   "FLX": "N41"
  }
 ],
\end{lstlisting}

The example presents the JSON object corresponding to the \textit{scuola} 'school'dictionary item. The semantic tags are stored in a JSON object containing all potential tagging. In this case, the word \textit{scuola} acquired a \textit{Nedi} Nedi tag since it is a noun for a building. Still, it also received a \textit{Num} tag because it can refer to ”all students and teaching and non-teaching staff who participate directly or indirectly in school activities”.

\begin{table}[h]
\caption{List of Concrete Nouns Tags with examples}
\label{tab:1}
\begin{tabular}{@{}lll@{}}
\toprule
\textbf{Tag} & \textbf{Description} & \textbf{Examples} \\
\midrule
Npc & body parts & leg, eye, tail\\
Npcorg & internal organs & heart, cell, neuron\\
Ntesti & texts & book, software, song\\
Nindu & clothings & jacket, shoes, glasses\\
Ncos & cosmetics & fard, eyeliner, shampoo\\
Ncibo & food & bread, chicken, meat\\
Nliq & not-drinkable liquids & gasoline, bleach, ammonia\\
Nliqbev & drinkable liquids & beer, wine, water\\
Nmon & money & euro, dollar, cent\\
Nedi & constructions & palace, wall, column\\
Nloc & locative & mountain, lake, town\\
Nmat & materials & mud, atom, iron\\
Nbot & plant life forms & tree, leaf, pine\\
Nfarm & medicines & aspirin, ibuprofen\\
Ndroghe & drugs & cigarettes, cocaine\\
Nchim & chemical elements & hydrogen, oxygen, carbonate\\
Ndisp & electronic devices & smartphone, computer, television\\
Nvei & vehicles & car, truck, muffler \\
Narr & furniture & closet, table, bed \\
Nstr & tools & fork, comb, stick \\
\botrule
\end{tabular}
\end{table}

Once we collected the dictionary resources, we performed a selection of them. Initially, we split the dictionaries into sub-lists based on semantic traits and concrete sub-tags. We select resources based on what we require for the experiments described in the following sections. This selection has been made by respecting a dimension criterion since matrices with too few dimensions can be too sparse, resulting in unpredictable similarity results. For this reason, we must choose only dictionaries containing several words greater than a predetermined limit. This limit was set at 200 dimensions. Smaller dictionaries have been merged with semantically similar resources when possible: the classes \textit{Npc} 'body parts' and \textit{Npcorg} 'internal body parts' and the classes \textit{Nliq} 'liquids' and \textit{Nliqbev} 'drinkable liquids' have been merged because they contain semantically similar elements. For the class \textit{Ncos}, there are no semantically related resources, and it has been discarded.

The final list of dictionary resources chosen includes 17 dictionaries from the Concrete Nouns dictionary.

We also merged \textit{Nloc} class with Toponyms. Finally, we add the dictionaries of Animal Nouns and Human Nouns to the final list of 17 dictionaries, bringing the list of nominal resources to 19.

Abstract Nouns deserve a separate discussion. They have been extensively studied in linguistics, although there is no unanimous description of this category \citep{khokhlova2014understanding}. Many authors underline the difficulty of sub-classifying Abstract nouns  \citep{Grochocka,zamparelli2020countability,husic2020vagueness}. For our work, some classes of abstract nouns can be beneficial, especially concerning the experiment we will propose about animal nouns. In particular, we have considered the idea of adding a dictionary of Colour nouns and a dictionary of Emotions. However, the number of elements of such resources will not overcome the 200 elements, and we finally decided not to generate matrices from abstract nouns. We used the verb matrices to capture abstract semantic properties based on the resources presented in the following sections.

\subsection{Semantic Predicates}
\label{predic}

As we said, LG Tables are organised in classes with syntactic criteria. Sixty-eight syntactic classes on almost 5000 verbs and 300 combinatory properties have been recognised for the Italian language. Contrariwise, the semantic classes of predicates are strongly separated from syntax and depend on the lexicon \citep{elia2013lexical}.\\ 
\cite{elia2010lexis} analysed the syntactic classes, searching for a convergence. In some classes, it is possible to find a generic low intersection, while other classes present a substantial semantic divergence. There are also classes which show a more robust convergence. Nevertheless, groups of 2-3 verbs per class have similar semantic properties. \cite{eliaLEG} detected 13 macro-semantic classes on more than 2000 Italian verbs. Among the semantic classes, Elia includes \textit{psychological verbs} ('to love', 'to fear'), \textit{body verbs} ('to kiss', 'to comb'), \textit{creation verbs} ('to create', 'to paint'), \textit{epistemic verbs} ('to know', 'to deduct'), \textit{personal transfer verbs} ('to give', 'to steal'), \textit{communication transfer verbs} ('to say', 'to extort') and \textit{locative transfer verbs} ('to enter', 'to leave', 'to maintain', 'to launch').

Those classes represent a first attempt at classification, are not exhaustive, and must be matched by other resources. Nonetheless, even in more recent research, it isn't easy to discover such an extensive lexical resource (particularly for verbs) constructed with a scientific and reproducible syntactic criterion. For this reason, we selected the LG classification as a base for our collection of verbs.

We started our collection of resources by considering two classes of verbs that can be used in an animal noun classification and are sufficiently robust to generate a functional distributional matrix. Those classes are \textit{Psychological}, \textit{locative verbs}. In addition, we defined another class which do not exist in the previous classifications. We named this class "\textit{verbs involving animated bodies}".

\subsubsection{Psych verbs}
Psych verbs are used to describe the emotional description of an animal. This description is subjective in many cases and is not part of an ontological classification of an animal concept, but it can capture common-sense knowledge about those concepts. For example, it is commonly accepted that a spider could cause terror or fear, and pandas or koalas cause tenderness. In those cases, they have the semantic role of the \textit{theme}. Nevertheless, animals can also be \textit{experiencer}. To achieve good performances with a DS matrix built over psych verbs, we need a model that considers the position of the arguments in the context of the verb. In fact, following the classification made by \cite{Belletti1988PsychverbsA}, those verbs have been classified into three classes which share the same thematic grid but have a different syntax: the subject of the verbs of class 1 represents the \textit{experiencer} and the object represents the \textit{theme} (\textit{temere} 'to fear', \textit{Antonio teme i ragni} 'Antonio fears spiders'), the subject of class 2 verbs represents the \textit{theme} (\textit{preoccupare} 'to worry', \textit{l'esame preoccupava Antonio} 'the exam worried Antonio') while third class includes intransitive verbs in which the subject is the \textit{theme} and the \textit{experiencer} is expressed by a prepositional complement (\textit{piacere} 'to like', \textit{i gatti piacciono ad Antonio} 'Antonio likes cats').

In our work, the syntactic Distributional Semantics (DS) model we used did not initially distinguish between argument positions. Given that the semantics of arguments for psychological and motion verbs change based on their syntactic roles, we adapted the original algorithm to account for this aspect in our research. We will detail the proposed modifications in subsequent sections. Concerning the Psychological verbs, we proceeded by selecting the complete list of psych verbs from the LG classification, which includes 630 verbs.

\subsubsection{Locative verbs}
The LG classification recognises 640 Locative predicates, classified into eleven semantic sub-classes. Among those sub-classes, we found verbs of movement, statics and scenic verbs, causative verbs and personal transfer verbs. The subject of those verbs is always an Agent/Cause (in \textit{Max attraversa la strada} 'Max cross the road', Max is \textit{Agent of motion}, while in \textit{Max sposta la sedia nella stanza} 'Max move the chair into the room' he is \textit{Cause of the motion action}). The role of Location is generally associated with the prepositional complement  (\textit{Antonio mette la penna \textbf{sul tavolo}} 'Antonio put the pen \textbf{on the table}'). Still, transitive verbs can present a crossed structure in which the \textit{Location} is the direct object (\textit{Antonio riempie \textbf{il camion} di casse} 'Antonio fills \textbf{the truck} with crates'), or a double structure (\textit{Antonio carica \textbf{il camion} di casse} 'Antonio loads \textbf{the truck} with crates', \textit{Antonio carica le casse \textbf{sul camion}} 'Antonio loads the crates \textbf{onto the truck}').

Transfer verbs are often sub-classified in different ways: \cite{nam1995semantics} proposes the following five semantic classes, based on the focus of the transfer event: a) goal locative; b) source locative; c) symmetric path locatives; d) directional locatives and e) stative locatives.\\
\cite{levin1993english} classified the Verbs of Motion into seven main classes, including \textit{verbs of inherently directed motion}, \textit{manner of motion}, \textit{motion using vehicles}. Other transfer verbs can be found among the \textit{verbs of putting} and \textit{verbs of removing} classes, \textit{verbs of possessional deprivation} classes, \textit{verbs of sending and carring} or \textit{verbs of possession} classes.

\cite{vietri2020lexicon} has deepened the LG classification by working on transitive motion verbs. \citeauthor{vietri2020lexicon} proposed six classes of patterns which correspond to a) the \textit{Locatum-location pattern}, in which the prepositional complement performs the role of origin (source) or destination (goal), or two complements denote the source and the goal; b) the \textit{Location-locatum pattern}, in which the direct object denotes the goal or the source and the prepositional complement indicates the \textit{locatum}; c) Locative alternation, which includes verbs with both structures; d) Non-denominal One-complement Transitive verbs, in which the direct object might be the source, the goal, a passage through or a median through; e) \textit{denominal Locatum} and f) \textit{denominal Location} verbs, in which the \textit{Locatum} or the \textit{Located entity} are derived by the base noun.

In \cite{vietri2020manner} the author recognised 210 Manner of Motion verbs (MMV). Vietri identified three main classes based on the \cite{folli2001two} classification. 
Class A includes inergative atelic verbs like \textit{gironzolare} ’to wander’, whose prepositional complement is not the goal but the generic location of the motion event. Class B contains verbs that can be inergative (atelic) or unaccusative (telic), such as \textit{correre} ’to run’, which can be used with the auxiliary verb \textit{avere} ’to have’ in \textit{Antonio ha corso al parco} ’Antonio ran into the park’ (atelic), or with auxiliary verb \textit{essere} ’to be’ in \textit{Antonio è corso al parco} ’Antonio ran to the park’; Class C includes unaccusative directional verbs (telic) such as \textit{avventurarsi} ’to venture’.

We unify the three classifications by selecting the transitive verbs classified by  \cite{vietri2020lexicon}, the MMV \citep{vietri2020manner}, but also the intransitive class of verbs classified by \cite{eliaLEG}, which includes the classes 20L - Journey (\textit{la bici attraversò il ponte} 'the bike crossed the bridge'), 7D - Destination (\textit{Max tornò al parco} 'Max returned to the park'), 7DP - Destination/Origin (\textit{il treno viaggia da Milano a Torino} 'The train travels from Milano to Torino'), 7S - Scenic (\textit{La nave naufraga nel lago} 'the ship wrecks in the lake'), 8ST - Static (\textit{Antonio abita a Venezia} 'Antonio lives in Venezia'). 

The total number of Transfer Verbs collected in our resource is 679, which includes intransitive (212), transitive (263, excluding denominal verbs), telic and atelic MMV (204). Since many verbs are homographs, the final number of selected verbs is 553.

\subsubsection{Verbs involving animated bodies}
Concerning the class of verbs involving animated bodies, we did not find a similar description in other predicate classifications. In the LG classification, for example, there are 55 \textit{body verbs} characterised by a double structure: they have a transitive structure with a human object which is correlated to a structure which expresses both the body part and the body (\textit{Antonio bacia Maria/Antonio bacia la guancia di Maria/Antonio bacia Maria sulla guancia} 'Antonio kisses Maria/Antonio kisses Maria's cheek/Antonio kisses Mary on the cheek).

Levin proposed a class of verbs \textit{involving the body} \citep{levin1993english}. The Levin classification includes eight classes, 3 of which are further subdivided. Levin’s classification includes verbs of bodily processes (snore, breathe, cry, inhale), verbs of nonverbal expression (cough, smile), verbs of gesture (blink, flutter (eyelashes), genuflect), Snooze and Flinch verbs, verbs of body-internal state of existence (convulse), suffocate verbs and verbs of bodily state and damage (pain, ache, burn, blanch). Those verbs can be found in many LG classes. Still, they share a consistent number of verbs with six (2, 2A, 2B, 10, 12, 20R), particularly twenty with the classes 2A (intransitive verbs with an animated subject such as \textit{sbadigliare} 'yawn'). The other classes are predominantly intransitive and have, in many cases, restricted distribution of a complement (2, 2B, 12 and 20R).\\

Starting from Levin’s classification, we created a new class of verbs that ”involves an animate being’s body”. The selection of those verbs is needed to complete the information about the animal domain by searching co-occurrences of terms with verbs that can select specific animals (for example, \textit{ruggire} 'to roar', but also \textit{partorire} 'to give birth' or \textit{deporre} 'to lay (eggs)'). The starting point for our selection is the Levin classification, but we used the LG tables as a resource to select Italian verbs and expand our list. LG Tables provide a large number of elements filled with distributional properties. Since we choose those verbs to build a Distributional Space, we rely on those properties to determine the predicates we defined as Animated Body verbs. In particular, we selected verbs with at least one argument restricted to animated nouns but not limited to human nouns.

We selected the following structures:

\begin{itemize}
    \item Verbs of Physical Process: this class includes verbs in which the subject is an animated being. Those verbs have an $N_0 V$ syntactic structure and are often derived from nouns, indicating the process's result. Those verbs alternate with a structure of the kind $N_0 releases V-N$. The only exception is the verb \textit{lacrimare} 'to tear', which also accepts a $N_0pc V Prep N_1$ structure in which the subject is a body part (\textit{gli occhi lacrimano ad Antonio/gli occhi di Antonio lacrimano/Antonio lacrima} 'the eyes tear to Antonio/Antonio's eyes tear/Antonio tears'). The verbs in this class are ten and include \textit{sanguinare} ’to bleed’, \textit{vomitare} ’to puke’.
    
    \item Verbs of Air emission: those verbs are similar to those of the previous class but do not indicate a physical process with a physical result. They could be considered ’manner of breathing’ verbs and include nine elements such as the verb \textit{russare} ’to snore’, \textit{respirare} ’to breathe’, or \textit{rantolare} ’to wheeze’.
    
    \item Verbs of Sound made by animals: we collected 17 verbs that indicate animal sounds. They have an $N_0 V$ structure with a restricted subject.
    
    \item Verbs of Physical state or change: that class includes verbs that indicate a physical condition or a change in a physical state. The structures of the verbs are $N_0 V$, in which the subject is an animated being. Semantically, those verbs indicate a process which results in a different state of existence. An example is \textit{addormentarsi} ’to fall asleep’, which suggests the passage between the awake state and the state of sleep. Forty-three verbs belong to this class, including \textit{morire} ’to die’, \textit{nascere} ’to born’ or \textit{ammalarsi} ’to get sick’.
    
    \item Specific Movement verbs: this class includes verbs of movement and manner of motion in which the subject can be an animal noun and, in some cases, which have a subject restricted to a subclass of animals. It belongs to those verbs \textit{galoppare} ’to gallop’, which is limited to horses and other equines, \textit{volare} ’to fly’, but also \textit{zoppicare} ’to limp’ or \textit{oscillare} ’to swing’.
    
    \item Verbs of Physical and sensory properties: 21 verbs which indicate a physical property such as \textit{profumare} ’to smell’ or \textit{puzzare} ’to stink’ where the subject is the beneficiary of the property or a sensory property such as \textit{prudere} ’to itch’ or \textit{dolere} ’to ache’ where the prepositional complement expresses the \textit{patient}.
    
    \item Verbs of Physical actions: 40 verbs that present different syntactic structures, each characterised by an animated subject. All those verbs represent an action started by the subject or involving the subject. For example, they belong to this class, the verbs \textit{reagire} ’to react’ or \textit{obbedire} ’to obey’ that accept an objective clause, and the verbs \textit{prevalere} ’to prevail’ and \textit{interferire} ’to interfere’, which are intransitive.
    
    \item Verbs of Behaviours: 63 transitive verbs accept a noun clause as a prepositional complement in this class. Those verbs select an animated noun as a subject or object. Among these verbs, we can find \textit{impegnare} ’to engage’, \textit{condurre} ’to lead’ or \textit{soprendere} ’to surprise’.
    
    \item Verbs of Social or Physical Interaction: this class of verbs includes 201 items which indicate a physical or social interaction between two animals or a human subject and an animal. The syntactic structures of those verbs include intransitive verbs with an $N_0 V Prep N_1$ structure in which there is a symmetric relation between $N_0$ and $N_1$ (\textit{il gatto si azzuffò con il cane/il cane si azzuffò con il gatto/il cane e il gatto si azzuffarono} 'the cat fought with the dog/the dog fought with the cat/the cat and the dog fought'); transitive verbs with a noun clause introduced by the preposition ’a’; transitive verbs with a noun clause that could be introduced by the preposition ’di’  (\textit{minacciare} 'to threaten', \textit{perdonare} 'to forgive'); transitive verbs with a noun clause introduced by the preposition 'da' (\textit{liberare} 'to free', \textit{guarire} 'to heal'). 
\end{itemize}

Table \ref{classiSEMSYN} includes a resumed description of those classes.

\begin{table}[h]
\caption{Quantitative description of the verbs involving animal bodies subclasses. }
\label{classiSEMSYN}
\begin{tabular}{@{}lll@{}}
\toprule
\textbf{Subclass} & \textbf{N. of Items} & Syntactic Structures \\
\midrule
Physical process & 10 & $N_0 V$\\
Air emission & 9 & $N_0 V$\\
Sound made by animals & 17 & $N_0 V$\\
Physical state of change & 43 & $N_0 V$\\
Movement & 19 & $N_0 V Prep N_1$\\
Physical and sensory properties & 21 & $N_0 V (di N_1)$\\
Physical actions & 40 & $N_0 V [Che F | Prep N_1]$\\
Behaviours & 63 & $N_0 V N_1 a Che F$\\
Social or physical interaction & 201 & $N_0 V [Prep N_1 | N_1 a Che F]$\\
\botrule
\end{tabular}
\end{table}

The final list contains 423 verbal uses, some of which may be homonyms. The term will only be included once because the distributional model we utilise to produce the matrices does not distinguish between homonymous words. There are 405 verbs in the list without repeated verbs. The ”AnimBody” dictionary includes all of those predicates. The following section explains how we use the lexical resources to build the Distributional resource.

\subsection{Construction of Matrices dataset}
\label{matrices}
As the first step of our process, we build a set of matrices with the same number of vectors, corresponding to the 17.000 Italian words with higher frequency values of the Paisà corpus \citep{lyding2014paisa}. 

To build the matrices, we use a syntactic algorithm called Syntactic Distance as Word Window (SD-W2) \citep{maisto2022extract} that analyses parsed texts and assigns a co-occurrence score based on the syntactic distance among sentence’s words: the Syntactic Distance is defined as the number of arcs of the dependency graph which separate two words. We used SD-W2 because we believe that a syntactic model could represent the co-occurrence relations in a better way despite the size of the corpus analysed. Our vectors consider as context only nouns, verbs or adjectives. The hierarchical representation of dependency graphs reduces the influence in the context of elements like prepositions and determiners, resulting in closer relationships between those elements in the sentences. In addition, the SD-W2 model allows us to choose base-map dictionaries, both vectors (the rows of the matrix) and dimensions (the columns), giving us more control over the matrix training procedure.

SD-W2 requires two parameters: the length of the syntactic window and the weighting function utilised. For this work, we use a window size of 2 and a GRAV weighting function, which assigns a score to the word corresponding to its POS Tag, favouring tags more interconnected in the syntactic schema (such as Nouns or Verbs).

We created a set of matrices for the experiment, each using one of the lists of words (nouns and verbs) selected and described in the previous section as columns. This means that each word is associated with a vector generated by calculating the syntactic co-occurrence value between the word and the words belonging to a specific semantic field or domain. For example, the matrix defined Animal contains 17.000 vectors corresponding to the most frequent words of the Paisà corpus and 2051 dimensions corresponding to the nouns of the Animal Noun Dictionary. Each vector dimension represents the co-occurrence between a generic word and an animal noun. The complete list of the produced matrices is shown in Table \ref{tab:2}.

\begin{table}[h]
\footnotesize
\caption{List of Matrices associated with the domain lexical resources}
\label{tab:2}
\begin{tabular}{@{}llll@{}}
\toprule
\textbf{Matrix} & \textbf{Lexical Resource} & \textbf{Description} & \textbf{Dimensions} \\
\midrule
GENERIC & most frequent words & generic matrix & 17.074\\
ANIMAL & Anl & Animals & 2051\\
BODILY & AnimBody & verbs involving animated bodies & 405\\
BODY & Npc + Npcorg & body parts and organs & 1206\\
BOTANIC & Nbot & plants and vegetables and their parts & 2125\\
BUILDINGS & Nedi & Buildings and their parts & 1567\\
CHEMISTRY & Nchim & chemical compounds and molecules &  1202\\
DEVICES & Ndisp & electronic devices & 1177\\
PHARMACY & Ndroghe + Nfarm & drugs and substances & 465\\
FOOD & Ncibo & food & 1148\\
FURNITURE & Narr & furniture & 445\\
HUMAN & Num & nouns referred to groups of persons & 11990\\
CLOTHING & Nindu & clothing & 982\\
LIQUIDS & Nliq + Nliqbev & any drinkable and undrinkable liquids & 359\\
LOCATIONS & Nloc + Toponyms & toponyms and location nouns & 2106\\
MATERIALS & Nmat & textiles or other materials & 1696\\
MONEY & Nmon & nouns of currencies & 216\\
PSYCH & Psych Verbs & psychological verbs & 630\\
TEXTS & Ntesti & any kind of texts & 1152\\
TOOLS & Nstr & mechanical tools & 3637\\
TRANSFER & Transfer Verbs & locative and movement verbs & 553\\
VEHICLES & Nvei & vehicles and their parts & 1071\\
\botrule
\end{tabular}
\end{table}

As mentioned in section \ref{predic}, the objectives of our resource necessitated differentiating the functions of verb arguments. Among the three classes of verbs we analyzed, we observed that the semantic function of an argument could vary depending on its syntactic position. For example, in the sentence 'rabbits fear foxes,' the SD-W2 model would assign the same co-occurrence value to \textit{rabbit} and \textit{fox} in relation to the verb 'to fear.' However, the semantic relationship between these nouns and the verb is, in fact, the opposite. \\
Taking this into account, we modified the original SD-W2 algorithm to discern among three distinct syntactic functions of the arguments: subject, object, and other. To facilitate this, we stored each verb from the dictionaries in the base map as three separate entities. Consequently, when the algorithm computes the co-occurrence between a word (e.g., rabbit) and the verb, it examines the dependency descriptor of the relationship between the noun and the verb and stores the value in the corresponding column (e.g., the subject of 'fear' for rabbit, the object of 'fear' for fox). By adopting this approach, the co-occurrence values for the two nouns with the same verb will differ. This technique was applied exclusively to the three dictionaries of verbs. As a result of this algorithmic adjustment, similarity scores between words that typically appear as the subject or object of a psychological verb could shift. To estimate these changes, we extracted the similarity values of various animal nouns before and after implementing the algorithmic modification. The findings revealed that a word like \textit{coniglio} (rabbit) shows lower similarities with the word \textit{topo} (mouse) by -5.4\% or \textit{volpe} (fox) by -4\%, and increased similarity with \textit{capra} (goat) by +7\% or \textit{lucertola} (lizard) by +5\%.

Each domain-specific matrix stands for a collection of constrained features that can more fully convey the meaning of a single word. From this list, we choose a subset of matrices that can accurately represent the characteristics of animal concepts to explore the concepts related to the macro-semantic class of animals for our experiments.
As an example of how the dimensions will influence the final similarity score, we tested GENERIC and HUMAN matrices on an animal ambiguous noun. We extracted the 20 most associated terms from the two matrices and then selected the five most associated for each 20. The result is an undirected graph in which the nodes are words, and the edges are the similarity scores between the two words they connect. We used Gephi \citep{bastian2009gephi} to represent the graphs, using the Modularity algorithm \citep{blondel2008fast} in order to identify classes of words based on the strength of their connections.

\begin{figure}[h]%
    \begin{subfigure}[t]{.5\linewidth}
    \centering\includegraphics[width=\linewidth]{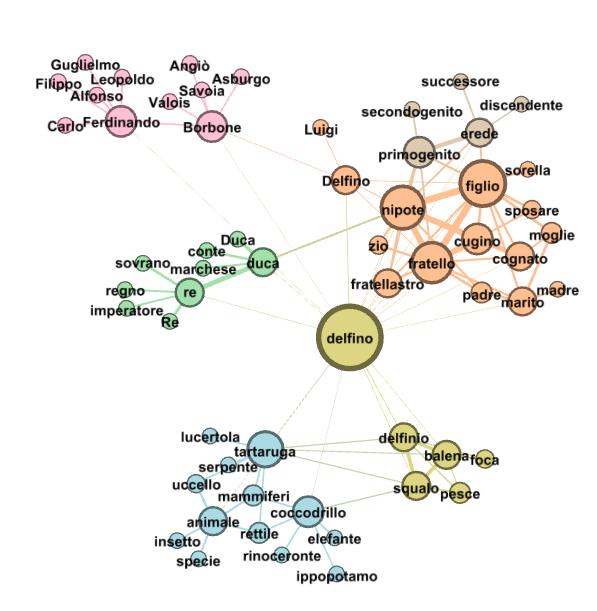}%
    \caption{GENERIC Matrix}
    \end{subfigure}
    \begin{subfigure}[t]{.5\linewidth}
    \centering\includegraphics[width=\linewidth]{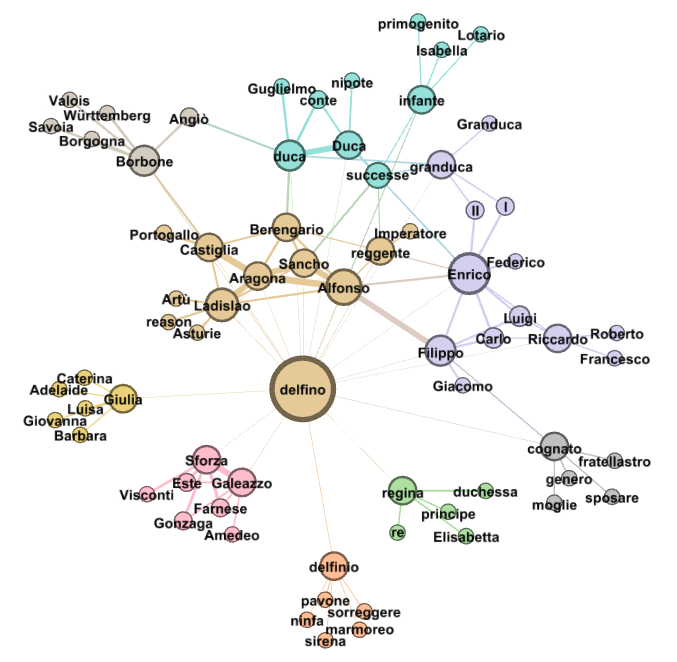}
    \caption{HUMAN Matrix}
    \end{subfigure}    
    \caption{Comparison between the semantic neighbourhood of the noun \textit{Delfino} 'Dolphin' generated from GENERIC and HUMAN Matrix}%
    \label{compa}%
\end{figure}

Figure \ref{compa} shows the differences between the two matrices. The word \textit{delfino} ’dolphin’, in fact, is an ambiguous word which indicates a cetacean but also ”the title given to the heir apparent to the throne of France”. In Figure  \ref{compa}(a), generated from the GENERIC matrix, the ambiguity is evident since we have two classes of words (on the bottom), which include animal nouns  (\textit{balena} 'whale' or \textit{tartaruga} 'turtles'), but also four classes whose words are directly connected with the second meaning of the word (\textit{re} 'king', \textit{figlio} 'son'). Contrariwise, in Figure \ref{compa}(b), animal nouns are almost absent (the noun \textit{pavone} 'peacock' is related to the noun \textit{delfinio} 'Delphinium', a plant, or 'Delphínion', an ancient Greek building, which probability is associated with \textit{Delfino} because of typos in the corpus\footnote{\textit{delfino} has 871 occurrences in Paisà while \textit{delfinio} has 772. In the IT-WAC corpus, the same word has respectively 12676 and 1}).

In the following sections, we present an experiment in which those matrices have been selected and combined to perform two semantics-related tasks, particularly with the semantics of animal nouns.

\subsection{DoMa User Interface}
We created a User Interface to make the use of the matrices easier. DoMa (DOmain MAtrices) software has been built in Java and is freely downloadable at https://figshare.com/ndownloader/articles/23898828/versions/3. The software allows users to easily select the matrices and extract similarities, neighbours or word vectors.

\begin{figure}
\centering
\includegraphics[width=0.7\textwidth]{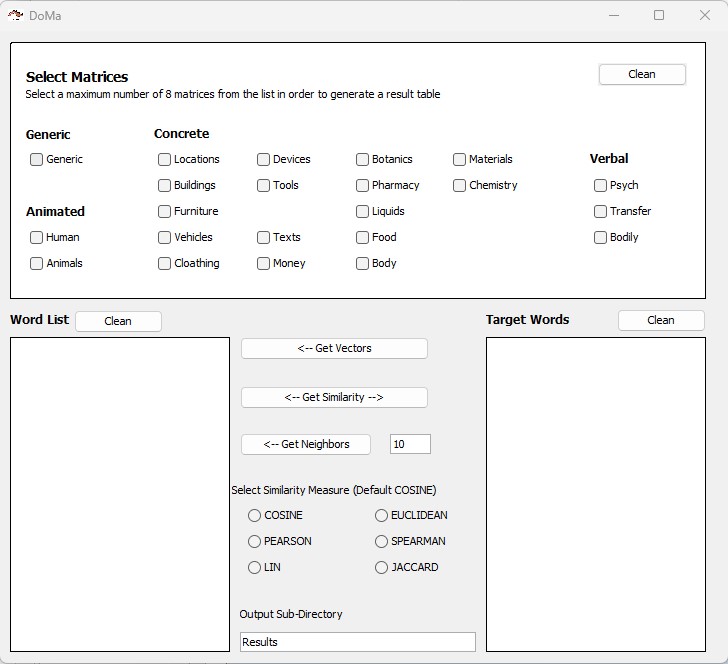}
\caption{The DoMa User Interface}
\label{doma}
\end{figure}

As shown in figure \ref{doma}, the software provides the set of matrices (including a generic matrix built on a dictionary of 18.000 most frequent words from Paisà corpus). Once the matrices are selected, the software searches simple words in the box on the left. The options are three:
\begin{enumerate}
    \item Get Vectors: the software generates a unique txt file in which each word in the ”Word List” box is associated with a matrix of vectors extracted from the selected matrices.
    \item Get Similarity: the software reads the words from the ”Word List” box and generates a \textit{.txt} file for each word. Each \textit{txt} file has a column for each word of the ”Target words” box and a value of similarity for each word and each selected matrix.
    \item Get Neighbors: the software generates a file for each word in the ”Word List” box. In each file, there is a list of the selected number of neighbours for each selected matrix.
\end{enumerate}

The software also allows users to select a specific similarity measure to perform the three tasks. Cosine Similarity is the default similarity matrix if a user does not specify any.

\section{Experimentation}
In this section, we present two experiments that can serve as helpful illustrations of two potential applications for our resource. The two experiments exploit two of the
main functions of \textit{DoMa}: vector and similarity extraction. 
The two experiments deal with

\begin{itemize}
    \item the categorising of animal nouns by combining vectors of the same nouns extracted from different matrices;
    \item the automatic extraction of the semantic characteristics of animals based on the semantic similarities between an animal and a prototype. 
\end{itemize}

In the next section, we will detail the two experiments and present the results of the two proposed methodologies.

\subsection{Automatic Classification of Animal nouns}
\label{classificationTask}
For this experiment, we relied on the idea that the matrices can be unified in a tensor to calculate the general similarity between two concepts characterised by the same set of features. We took inspiration from Compositional Distributional Semantics \citep{baroni2014frege} and the works of \cite{mitchell2010composition} that compare different compositional metrics. This kind of model is based on the idea that the meaning of a sentence or a phrase can be analysed as a result of a composition of the co-occurrence vectors of the words in it. \cite{mitchell2010composition,mitchell2008vector} studied the behaviour of different composition operations on small phrase examples, such as the sequences ”old dog” and ”old cat”. Even though the vector addition operation is the most frequently used, they disregard it because the results can occasionally be affected by the co-occurrence values of a single word. With vector products, \citeauthor{mitchell2010composition} achieve better results. 

We hypothesise that, similarly to the meaning of a sentence or a phrase, the meaning of a single word can be constructed starting from a set of matrices that describe a particular feature. Instead of combining vectors of the words composing a sentence or a phrase, we extract vectors from different domain matrices for the same animal noun. The final goal of this experiment is to build a network of similarities among animal nouns to calculate sub-classes automatically. We also produce the same network using single matrix similarity values to compare the effectiveness of our methodology.

We cannot use vector addition or vector product as a composition function, as \cite{mitchell2010composition} suggested because the dimension of a vector extracted from a matrix does not correspond with the dimension of the vectors extracted from other matrices. The first dimensions of the ANIMAL matrix, for example, might be, in alphabetical order \textit{ameba} 'amoeba' \textit{antilope} 'antelope',  or \textit{aquila} 'eagle', since the first dimensions of the BODY matrix could be \textit{adduttore} 'adductor', \textit{alluce} 'toe' or \textit{avambraccio} 'forearm'.

\cite{baroni2010distributional} proposed to use tensors to collect syntactic tuples data of different types. They combined the distributional information of a set of weighted \textit{word-link-word} tuples into a third-order tensor. A tensor is a multidimensional array that describes a multilinear relationship between groups of objects (i.e. vectors). A vector represents a first-order tensor. Matrices are second-order tensors. Third-order tensors are arrays with three \textit{components}such as a syntactic relationship between two words (for example, \textit{the soldier has shot} is represented by a tuple \textit{<soldier,}sbj\_intr\textit{,shot} whose elements' vectors represent the components of the tensor). In their work, \cite{baroni2010distributional} build a single third-order tensor, extracting specific components from it in order to generate specialised matrices. Our approach is different: we start from many specialised matrices and aim to compose the tensor by selecting only the matrices that could effectively describe a concept. Since ”just like we can measure degrees of similarity of two or more vectors living in the same vector space, we can measure the similarity of matrices and higher-order tensors” \citep{baroni2014frege}, we generate a network of concepts by extracting the similarity between concept tensors built from the selected vectors of the corresponding noun.

In our experiment on the Animal macro-semantic class, we selected a set of matrices that can impact the classification of animal nouns. The selection of those matrices reflects some fundamental characteristics of the nouns that belong to the zoology domain:

\begin{itemize}
    \item ANIMALS, 2051 dimensions: it reflects a conceptual association among animals. For example, \textit{whale} and \textit{dolphin} are similar because they have words that denote fishes or other marine animals in their context.
    \item BODY, 1206 dimensions: this matrix produces similarities between nouns which appear near the same body part, such as \textit{whale} and \textit{dolphin}, frequently appearing with the noun \textit{fin}, \textit{tail}, etc.
    \item FOOD, 1148, dimensions: it relates animals which eat similar foods. For example, \textit{giraffe} and \textit{elefant} eat \textit{grass}, \textit{leaf} and other \textit{vegetables}.
    \item TRANSFER, 553 dimensions: this matrix includes verbs of motion and relates animals with a similar manner of motion or generally associated with the same motion event. \textit{Whales} and \textit{dolphins} \textit{swim}, and \textit{lions} and \textit{leopards} \textit{leap}, \textit{run} and \textit{climb} trees.
    \item LOCATION, 2016 dimensions: it relates animals which live in similar habitats. \textit{Lion}, \textit{leopard}, and \textit{elephant} have words such as \textit{savannah}, \textit{grassland}, \textit{zoo}, or \textit{Africa} in their contexts.
    \item PSYCH, 630 dimensions: this matrix is related to the emotion an animal can feel or cause. This information derives from common sense reasoning but could be helpful in our classification task. \textit{Spider}, for example, is often related to fear by the verb \textit{to scare} or \textit{to fear}.
    \item BODILY, 405 dimensions: the dimension of this matrix includes verbs related to the body and generates similarities between animals with similar natural behaviours. For example, an \textit{eagle} and a \textit{duck} are similar because both \textit{lay eggs}, \textit{tigers} and \textit{lions} because they \textit{roar}.
\end{itemize}

\begin{figure}[H]
\centering
\includegraphics[width=0.8\textwidth]{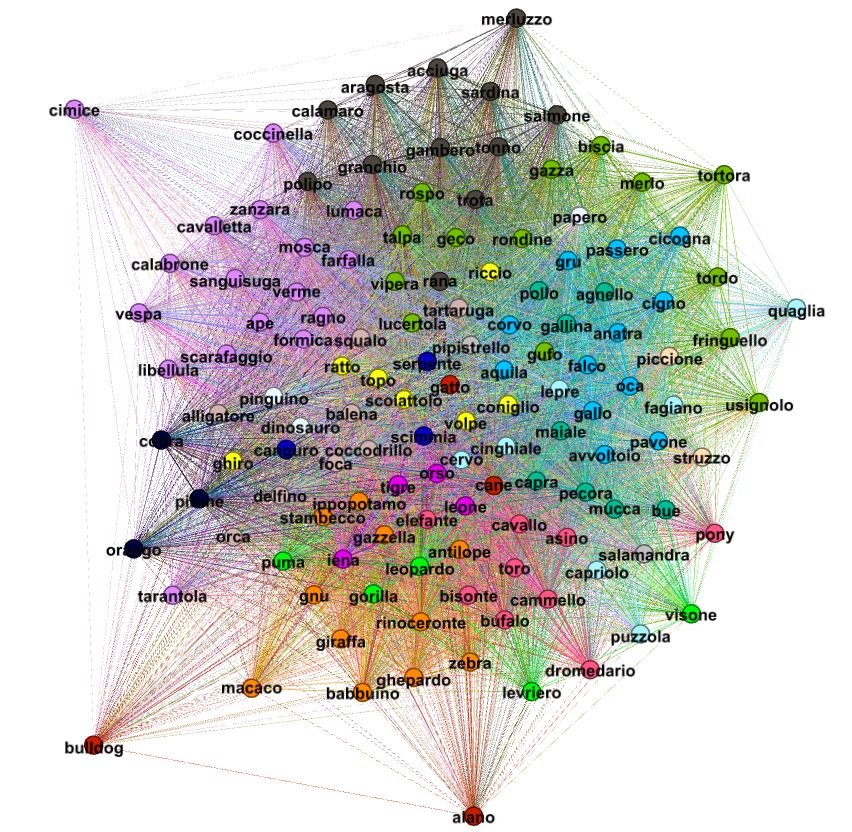}
\caption{The network of Animal Nouns}
\label{aninet}
\end{figure}

We extracted the seven vectors for 131 animal nouns, reducing the vector dimensions to 200 with Singular Value Decomposition \citep{rohde2002methods} to generate a quadratic matrix for each animal noun. Although the tensor must be composed of matrices with the same number of dimensions, each macro-semantic area represented by a matrix must contribute to the tensor with the same amount of information to the final similarity score.\\
The final tensor is then composed using the matrices generated for each animal noun.

We use TensorFlow \citep{abadi2016tensorflow}, a Python library with many state-of-art similarity measures, to build the tensors. We extracted similarity values between each pair of animal nouns with Cosine Similarity, generating a list of Noun-Noun-Similarity triples.

This list represents an undirected graph with weighted edges. We used Gephi to visualise the graph and perform a classification operation on those 131 nouns. Figure \ref{aninet} shows the graph.

The animal nouns have been classified through the \textit{Modularity} network classification algorithm  \citep{blondel2008fast}. The modularity class algorithm has been set to produce more classes (a Resolution \citep{lambiotte2008laplacian} score of 0.7) and generates 18 classes. In the graph, each class is distinguished by the node's colour. The position of each node is calculated by another network algorithm called Force Atlas 2, a linear-attraction linear-repulsion model provided by Gephi.

\begin{figure}[H]
    \begin{subfigure}[t]{.5\linewidth}
    \centering\includegraphics[width=\linewidth]{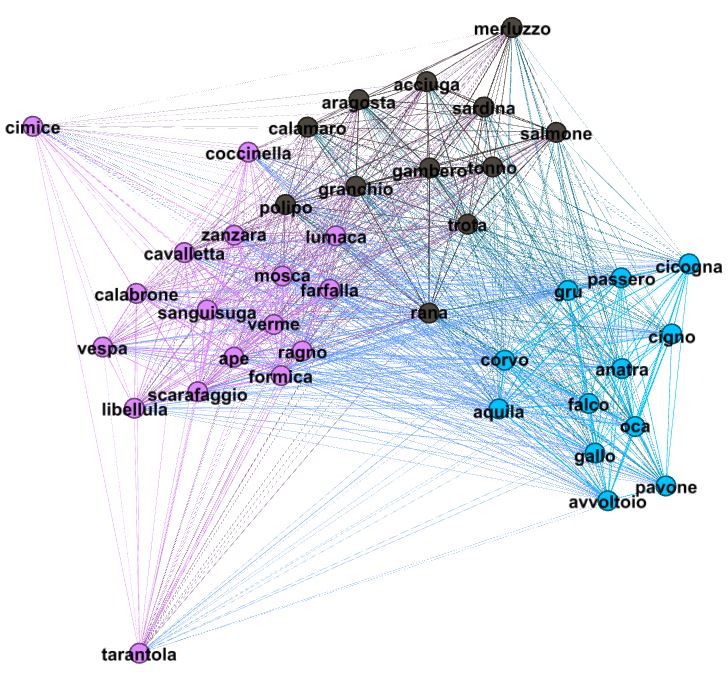}%
    \caption{Classes 0 (blue), 1 (black) and 14 (purple)}
    \end{subfigure}
    \begin{subfigure}[t]{.5\linewidth}
    \centering\includegraphics[width=\linewidth]{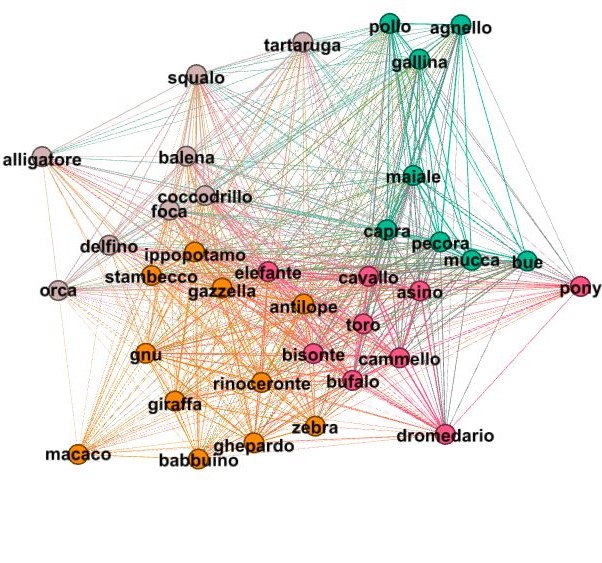}
    \caption{classes 4 (green), 8 (brown), 9 (fuchsia) and 16 (orange)}
    \end{subfigure}    
    \caption{A detailed vision of some classes generated by the Modularity class}%
    \label{fig2}%
\end{figure}

Figure \ref{fig2} shows a detailed view of two class groups denoting an incredible internal coherence. In particular, we can observe the classes of Figure \ref{fig2}(a), which include insects, spiders and invertebrates (class 14), birds (class 0), and sea animals (class 1). In Figure \ref{fig2}(b) we have classes of animals which represent non-taxonomic classes that we can still consider correct. We have a class of farm animals (class 4), a class of pack animals (class 9), African animals (class 16) and other animals which are not fishes but commonly live in water (class 8), such as \textit{balena} 'whale',  \textit{alligatore}, 'alligator', \textit{delfino}, 'dolphin' or \textit{tartaruga}, 'turtle'.\\
The graph in \ref{fig2}(b) is newsworthy because it demonstrates the influence of different matrices in generating similarity scores. For example, the animals in class 8 are similar because they have similar contexts concerning LOCATION (sea, water, river, etc.), but also relating to TRANSFER (to swim, to float). However, they are sufficiently distinct from fish to not belong to the same class. Class 4 can also be influenced by LOCATION and FOOD since they are all eatable animals. 

We executed the same analysis using the identical set of 131 animal nouns, comparing the similarity values derived from the Word2Vec and BERT algorithms. For Word2Vec, we utilized the Spacy Italian pipeline model (\textit{it\_core\_news\_lg} pre-trained model), which includes a matrix of 500,000 vectors, each with 300 dimensions. For BERT, we employed the \textit{bert-base-italian-xxl-cased} model. In the case of BERT, a preliminary step of fine-tuning was necessary to extract vectors for the selected animal nouns. We compiled a corpus by identifying every occurrence of each animal noun in Wikipedia and incorporating the entire article. Following the recommendations in \cite{chersoni2021decoding}, we also included Wikipedia articles that mention other meanings of the animal noun. Ultimately, we determined the similarity score for each noun by calculating the average similarity across the last four layers for all word senses of the nouns.   

\begin{figure}[H]
\centering
\includegraphics[width=0.8\linewidth]{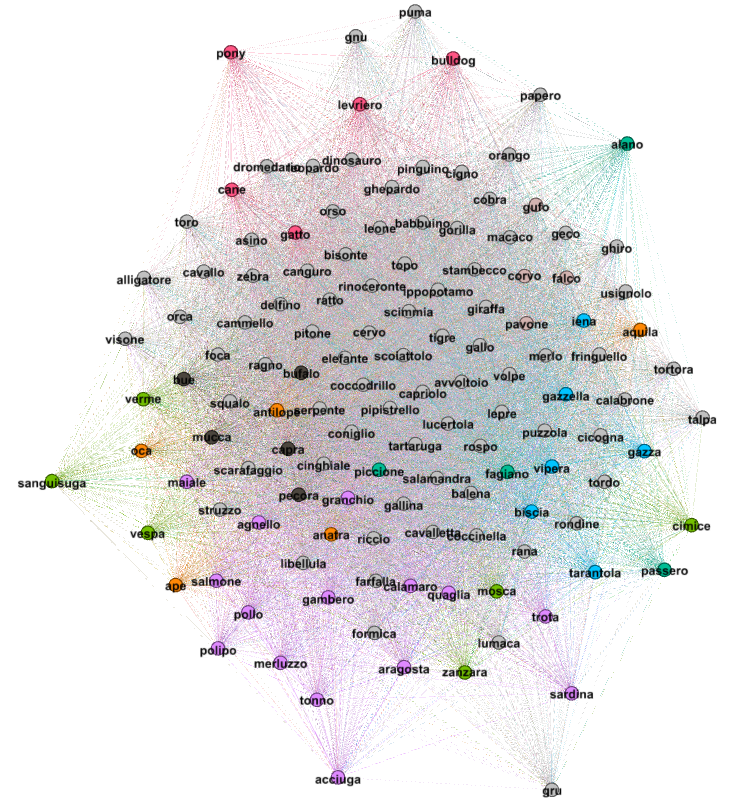}
\caption{General view of the network produced with word2Vec similarities}
\label{w2vgen}
\end{figure}

In Figure \ref{w2vgen}, we display the network of animals generated from the distances obtained through the W2V model. The graph produced by BERT was significantly scattered, leading us to decide against including it as a visual representation. The differences with the graph produced by the re-composition of feature matrices are evident at first look. We used the same parameters for the Modularity class and obtained a much more interconnected graph and many tiny classes (38). Furthermore, even if smaller, the classes often include animals that seem out of line with the class. For example, blue class includes two African animals (\textit{iena} 'hyena' and \textit{gazzella} 'gazelle'), but also snakes (\textit{vipera}, 'viper' and  \textit{biscia}, 'grass snake'), a spider (\textit{tarantola} 'tarantula') and a bird (\textit{gazza} 'magpie'). The results from BERT were even less satisfactory, as it produced 75 distinct classes, of which only one class contained three nouns (lamb, donkey, and eagle), while the majority of classes comprised only two nouns.

The unsatisfactory results obtained with BERT indicate potential issues with the Italian version of BERT that we utilized. For instance, the Italian tokenization process tends to morphologically split many words, complicating the task of extracting word vectors. Moreover, the corpus used for fine-tuning might have been too small, further impacting the quality of the results. In general, we noticed that the range of similarity scores extracted from BERT was limited, and therefore the differences between the animals were not as pronounced as to allow for correct classification.

To compare with a newer model, we submitted the classification task to ChatGPT (Version 4.0), providing it with the list of 131 animal nouns and requesting a classification based on general similarity. ChatGPT produced 5 classes, which contained various errors. In the first class, 'Terrestrial Mammals,' it incorrectly included 'gallina' ('chicken'), explaining that 'although primarily a bird, I include it here for domestic familiarity,' and 'trota' ('trout') was included 'by mistake.' Similar types of errors were observed in the other classes: 'birds' (which erroneously included dinosaurs), 'water animals' (which incorrectly included spider, bee, python, antelope, bison, etc.), and 'insects' (where 'vipera' was misplaced). The last class, 'reptiles and amphibians,' did not contain errors. 

Lastly, we conducted a quantitative evaluation of the classes by assessing the precision of the classification across the entire set of nouns. We asked four individuals to assign the most probable name to each class. We then deemed every noun that was semantically related in any way to the class as correctly classified. For instance, the yellow class from the SD-W2 classification was labelled as the 'rodents class,' and we considered the nouns \textit{riccio} ('hedgehog'), \textit{ratto} ('rat'), \textit{topo} ('mouse'), \textit{scoiattolo} ('squirrel'), \textit{ghiro} ('dormouse'), and \textit{coniglio} ('rabbit') as correctly included, but not \textit{volpe} ('fox'). The results of the quantitative evaluation are presented in Table \ref{classprec}.

\begin{table}[h]
    \centering
    \caption{Accuracy values for three tested models on the classification task.}
    \begin{tabular}{l|l|l}
    \textbf{Model} & \textbf{Precision} & \textbf{Number of Classes} \\
    \hline
       SD-W2  &  0.80 & 18\\
         W2V & 0.64 & 38\\
         GPT4 & 0.77 & 5 \\
    \end{tabular}
    \label{classprec}
\end{table}

The study of the results suggests that our method can improve noun classification through semantic similarities since it describes concepts of a specific sub-domain more precisely. Considering the differences in technology between the proposed method and GPT4, our method reaches a surprising result.
In the next experiment, we will use the matrices to automatically extract animal features from their similarity with a set of prototypical animals.

\subsection{Automatic Concepts Description}
This experiment uses domain matrices to create a list of distinguishing features for an animal. This is achieved by comparing the animal to a group of prototype animals and identifying similarities.

The idea is to connect semantic features and prototype animal (PA) nouns. Each feature will be associated with a list of PAs. For example, the words \textit{bee}, \textit{butterfly}, and \textit{eagle} are associated with a list of features describing their physical properties or behaviours. Some of those features can be \textit{fly}, \textit{is\_an\_insect}, \textit{has\_poison} or \textit{is\_a\_bird}.\\
We must associate those properties to each prototype as follows:

\begin{itemize}
    \item \textbf{vola} 'fly': \textit{ape} 'bee', \textit{farfalla} 'butterfly', \textit{aquila} 'eagle'
    \item \textbf{è\_un\_insetto} 'is\_an\_insect': \textit{ape} 'bee', \textit{farfalla} 'butterfly'
    \item \textbf{è\_velenoso} 'has\_poison': \textit{ape} 'bee'
    \item \textbf{è\_un\_uccello} 'is\_a\_bird': \textit{aquila} 'eagle'
\end{itemize}

Depending on the type of feature we are considering, we choose a different domain matrix and extract similarity values between each non-prototypical animal (NPA) noun and PA noun. We calculate the similarity values between these animals and the prototypes in order to describe two NPA, such as \textit{calabrone} ’hornet’ and \textit{corvo} ’crow’. Then, we generate an index of relationships between the NPA and each feature. For example, the word \textit{calabrone} will have an average high value of similarity with \textit{ape}, a medium average value with \textit{farfalla} and a low value with \textit{aquila}; contrariwise, \textit{corvo} will have substantial average similarity with \textit{aquila} but not with \textit{ape} or \textit{farfalla} (table \ref{hornetCrow}).

In this study, we suggest a method for categorizing features of NPAs such as \textit{calabrone}, which possess attributes like \textit{è\_un\_insetto}, \textit{è\_velenoso} and \textit{vola}. Similarly, NPAs like \textit{corvo} can be classified based on their features \textit{è\_un\_uccello} and \textit{vola}. To achieve this, we link each feature to a specific matrix, such as ANIMAL for taxonomic classification (\textit{è\_un\_insetto}), BODY for body parts (\textit{ha\_il\_pungiglione} 'has\_sting' or \textit{ha\_le\_ali} 'has\_wings') and TRANSFER for locomotion (\textit{vola} 'fly'). By doing so, we can generate more precise values of similarity. Finally, we establish a threshold to determine when a property can be attributed to the non-prototypical noun.

\begin{table}[h]
\caption{Similarity between NPA nouns \textit{calabrone} 'hornet' and \textit{corvo} 'crow' and PA nouns \textit{ape} 'bee', \textit{farfalla} 'butterfly' and \textit{aquila} 'eagle' in different matrices}
\label{hornetCrow}
\begin{tabular}{@{}llll@{}}
\toprule
Nouns & \textbf{ANIMAL}& \textbf{BODY} & \textbf{LOCATIVE} \\
\midrule
\textbf{Calabrone} & & & \\
\textit{ape} & 0.47 & 0.03 & 0.2\\ 
\textit{farfalle} & 0.34 & 0.0 & 0.12\\
\textit{aquila} & 0.04 & 0.02 & 0.06\\
\hline
\textbf{Corvo} & & & \\
\textit{ape} & 0.04 & 0.02 & 0.28\\ 
\textit{farfalla} & 0.24 & 0.0 & 0.40\\
\textit{aquila} & 0.62 & 0.23 & 0.43\\
\botrule
\end{tabular}
\end{table}

To conduct this experiment, we select a set of features that could describe the prototypes and be easily connected to the feature matrices shown in section  \ref{matrices}, as well as semantic resources to identify the group of prototypes, i.e., a ”central” salient instance on which a semantic category is developed \citep{rosch1973natural}. We tested the model on the animal domain because it is easy to understand since it regards common-sense knowledge and has different semantic resources available \citep{mcrae2005semantic,lieto2017dual}. In addition, a solid base of nouns belonging to this domain appears many times in corpora. We recognised over 150 animal nouns in our corpus with an average of almost 1100 occurrences, from the 5-6000 of the noun \textit{gatto} ’cat’ or \textit{leone} ’lion’ to the 1-2000 of \textit{maiale} ’pig’ and \textit{orso} ’bear’, with only a small amount of very specific nouns with few occurrences (\textit{rottweiler} 23 occurrences, \textit{bonobo} 50 occurrences or \textit{anaconda} 52).

For the first task, recognising a set of PAs, we deal with DUAL PECCS  \citep{lieto2017dual}, a cognitive system for conceptual representation and categorisation. \citeauthor{lieto2017dual} built a list of prototypical animals with an accurate description involving physical properties but also location, locomotion and behaviour properties. The list of animals included in DUAL-PECCS has been selected for two main reasons:
\begin{itemize}
    \item Firstly, it provides an accurate description of a significant number of common animals, which are prototypical instances of a specific race, but also of a more general subdomain: the description of the concept of \textit{tiger} is prototypical for the tiger race (orange and black stripes, etc.), but also for the sub-domain of \textit{Asiatic felines} and, more in general, for the category of \textit{big cats}.
    \item Secondly, the animals cover an extensive range of exemplars easily distinguishable by physical description, location, behaviours and scientific categorisation.
\end{itemize}

DUAL PECCS descriptions are written in XML format and are characterised by the following structure:

\begin{figure}[H]
    \centering
    \includegraphics[width=8cm]{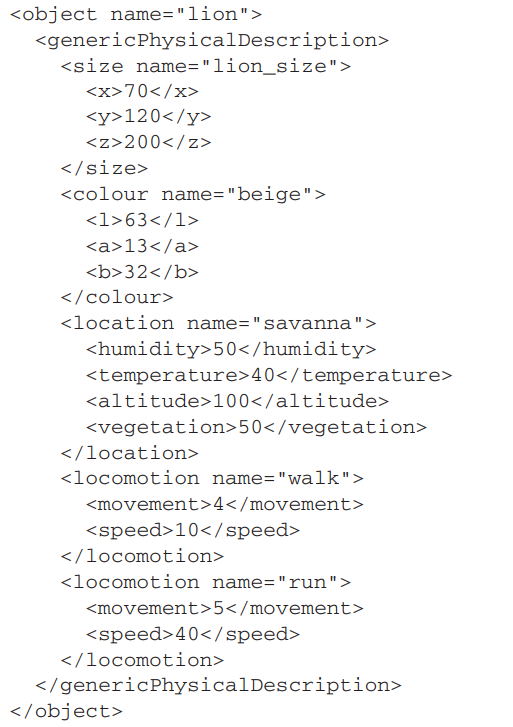}
    \caption{Example of "Lion" concept structure in DUAL PECCS from \cite{lieto2017dual}}
    \label{dual}
\end{figure}

For the collection of a set of features that can describe the prototypical animals, we rely on the work of \cite{mcrae2005semantic}, which described a large set of semantic feature production norms (2.526) for 541 living and nonliving concepts corresponding to a single noun. Students from three different universities have collected the features norms. Each participant received a form with a list of concepts and ten blank lines to list features. The features must be of different types, such as physical (perceptual) properties, functional properties and other facts (category or other encyclopedic facts). Then, all the similar feature norms were unified, and a concept vector was built for each concept, starting from the list of all the norms collected. An example of norms that describe the concept of \textit{duck} is:
\begin{itemize}
    \item \textbf{duck}
    \item[] is a bird
    \item[] is an animal
    \item[] waddles
    \item[] flies
    \item[] migrates
    \item[] lays eggs
    \item[] quacks
    \item[] swims
    \item[] has wings
    \item[] has a beack
    \item[] has webbed feat
    \item[] has feathers
    \item[] lives in ponds
    \item[] lives in water
    \item[] hunted by people
\end{itemize}

This description results from the union of all the feature norms cited by almost one participant. It contains references to different types of properties and results in an excellent abstract description of the concept \textit{duck}.

In our work, we automatically build a similar description of a list of animals that could not be considered prototypical and for what we have not a feature/property description.\\
In the next section, we illustrate the steps involved in our experiment in detail.

\subsubsection{Description of prototypes}
From DUAL-PECCS, we extract a list of 49 prototypes belonging to the macro-class of animals. From the original list, we excluded two nouns: \textit{lince} ’lynx’, because they appear with shallow frequency in the corpus used to train the matrices (243); \textit{pantera} ’panther’ because the list already includes other ”big cats” as \textit{lion}, \textit{tiger} and \textit{cheetah} that are more frequent in texts and provides a good features variation for that subclass of animals. The total number of prototypes collected was 47.

Regarding the features, we identified nine macro families of descriptive properties and 89 single features. The features have been extracted initially from the list of feature norms of \citeauthor{mcrae2005semantic}, but they have been integrated with other properties of the same kind that are not in McRae’s list. For example, in the original list, there is the property \textit{è\_un\_mammifero} 'is\_a\_mammal' or \textit{è\_un\_pesce} 'is\_a\_fish', but there are no references to ’amphibians’ or ’reptiles’, probably because there were no individuals from that classes in the list of concepts proposed by the authors to the participants. The nine feature families are associated with a specific feature matrix, as listed in Table \ref{tab:5}.

At this step, we have a list of prototypes described by 89 features divided into nine macro-families, each associated with one feature matrix. We need to associate each feature with a list of prototypes which can be described by it. The final list of prototypes is the following:

\begin{multicols}{3}
    \begin{itemize}
    \scriptsize
\item[]	ape,\textit{	bee	}
\item[]	aquila,\textit{	eagle	}
\item[]	asino,\textit{	donkey	}
\item[]	avvoltoio,\textit{	vulture	}
\item[]	balena,\textit{	whale	}
\item[]	cammello,\textit{	camel	}
\item[]	canguro,\textit{	kangaroo	}
\item[]	cavallo,\textit{	horse	}
\item[]	cervo,\textit{	deer	}
\item[]	cigno,\textit{	swan	}
\item[]	coccinella,\textit{	ladybug	}
\item[]	coccodrillo,\textit{	crocodile	}
\item[]	coniglio,\textit{	rabbit	}
\item[]	delfino,\textit{	dolphin	}
\item[]	elefante,\textit{	elephant	}
\item[]	farfalla,\textit{	butterfly	}
\item[]	formica,\textit{	ant	}
\item[]	gallina,\textit{	hen	}
\item[]	gatto,\textit{	cat	}
\item[]	ghepardo,\textit{	cheetah	}
\item[]	giraffa,\textit{	giraffe	}
\item[]	granchio,\textit{	crab	}
\item[]	gufo,\textit{	owl	}
\item[]	ippopotamo,\textit{	hippopotamus	}
\item[]	koala,\textit{	koala	}
\item[]	leone,\textit{	lion	}
\item[]	lumaca,\textit{	snail	}
\item[]	merlo,\textit{	blackbird	}
\item[]	mucca,\textit{	cow	}
\item[]	orso,\textit{	bear	}
\item[]	pinguino,\textit{	penguin	}
\item[]	pipistrello,\textit{	bat	}
\item[]	pitone,\textit{	python	}
\item[]	polpo,\textit{	octopus	}
\item[]	ragno,\textit{	spider	}
\item[]	rana,\textit{	frog	}
\item[]	rinoceronte,\textit{	rhinoceros	}
\item[]	salamandra,\textit{	salamander	}
\item[]	salmone,\textit{	salmon	}
\item[]	scimmia,\textit{	monkey	}
\item[]	serpente,\textit{	snake	}
\item[]	squalo,\textit{	shark	}
\item[]	tartaruga,\textit{	turtle	}
\item[]	tigre,\textit{	tiger	}
\item[]	topo,\textit{	mouse	}
\item[]	toro,\textit{	bull	}
\item[]	volpe,\textit{	fox	}
    \end{itemize}
    \end{multicols}

\begin{table}[h]
\caption{Feature-Families associated with respective Matrices and an example of feature}
\label{tab:5}
\begin{tabular}{@{}lll@{}}
\toprule
\textbf{Family} & \textbf{Example of Feature} & \textbf{Feature-Matrix} \\
\midrule
Category & mammal, fish, reptile & ANIMALS\\
Body Surface & fur, plumage, scales & BODY\\
Locomotion & number of paws, wings, fins & BODY\\
Other body parts & tail, beak, sting & BODY\\
Food preferences & vegetables, meat, fish & FOOD\\
Climate & polar, tempered, tropical & LOCATION\\
Habitat & wood, savanna, ocean & LOCATION\\
Type of locomotion & fly, gallop, crawl & TRANSFER\\
Other behaviours & lay, bite, scratch & BODILY\\
\botrule
\end{tabular}
\end{table}

For some features, the list of associated prototypes is easy to generate. The features belonging to the family ’Category’ as \textit{è\_un\_mammifero} 'is\_a\_mammal' will include all the mammal exemplars of our list of PA and all these words can be associated only with one feature of the family. \\
For other families, such as ’Climate’, we select animals commonly associated with a specific feature and assign the same PA noun to more features. For example, the PA noun \textit{elefante} 'elephant' is typically associated with Africa, but also with Asia, so we connect the word \textit{elefante} with both features \textit{vive\_in\_Africa} 'live\_in\_Africa' and \textit{vive\_in\_Asia} 'live\_in\_Asia'. Nevertheless, we don't consider captive elephants, because their habitat is not prototypical.\\
We cannot include the PSYCH matrix in this classification because we didn’t have a psychological description of the prototypes. Another family of features we must discard is Colour because we didn’t generate a matrix over the colour dictionary since it is too small to work correctly.

\subsubsection{Assign features to non-prototypical animals}
To assign a description to an NPA, we must extract the similarity values between the animal and each prototype assigned to a feature for every selected matrix. We then transform and combine the similarity values in order to obtain an index of association between the animal and the feature.\\
It is necessary to transform the similarity values because various matrices produce them at very different rates. These values, in many cases, are not comparable to each other. In addition, we needed to smooth the similarity of feature matrices subject to excessive fluctuation of some values due to their lack of dimensionality. For this reason, we performed two kinds of transformation of the similarity values extracted from the involved matrices:

\begin{equation}
    Weighted Similarity: Wsim_{(p,t)} = \frac{2S_{(p,t)} + Sg_{(p,t)}}{3}
    \label{weiSim}
\end{equation}

\begin{equation}
    Percentage Similarity: Psim_{(p,t)}= \frac{Wsim_{(p,t)}}{\sum Wsim_{(p,t)}}
    \label{perSim}
\end{equation}

First, we generate the \textit{Weighted Similarity} (\ref{weiSim}) values by performing a weighted average between the similarity (S) of the target word (t) and the prototypes (p) obtained with a feature matrix ($S_{(p,t)}$), and the similarity obtained with the generic matrix ($Sg_{(p,t)}$). \\
Then we generate a \textit{Percentage Similarity} (\ref{perSim}) by calculating to what extent the similarity value between the animal and a prototype contributes to the sum of all the similarities between the same animal and every prototype for a specific feature matrix. The following step is to generate the index of association. We combine three values:

\begin{itemize}
    \item an average of similarity between the animal and the related prototypes ($S_{rel}$) (\ref{avrelsim});
    \item an average of similarity between the animal and all the unrelated prototypes ($S_{unrel}$) (\ref{avunrelsim});
    \item a measure of Centrality ($C_t$) (\ref{centr}) meant as the proportion between high related and all the related prototypes.
\end{itemize}

 \begin{equation}
     Average Feature Similarity: S_t = \frac{\sum Psim_{(p,t)}}{N_{p}}
     \label{avfeatsim}
 \end{equation}

 \begin{equation}
     Average Related Similarity: S_{rel} = S_t with p \in f
     \label{avrelsim}
 \end{equation}

  \begin{equation}
     Average Unrelated Similarity: S_{unrel} = S_t with p \ni f
     \label{avunrelsim}
 \end{equation}

  \begin{equation}
    Centrality: C_{t} = LOG_{N_p} 1 \times Psim_{(p,t)} > CK with p \in f
    \label{centr}
 \end{equation}

The \textit{centrality} of a target word for a feature is calculated as a logarithm of the number of prototypes with high values (more than the CK parameter) of $Psim$ with that word. The logarithm considers the total number of prototypes associated with the features as the base.\\
We combined the three values in an index of feature’s association $F_t$ calculated as follows:\\

  \begin{equation}
    F_t =  S_{rel} - S_{unrel} + C_{t}
    \label{FT}
 \end{equation}

Table \ref{tab:6} shows the similarity score obtained by the noun \textit{bue}, 'ox', with the nouns of the prototypes related to the features \textit{ha\_corna} 'has\_horns' and \textit{è\_velenoso} 'has\_poison', both belonging to the feature family ''other body parts'' and with the feature-matrix BODY.

\begin{table}[h]
\caption{Similarity values and Percentage Similarity Values between the word \textit{bue}, 'ox', and prototypes related with the features \textit{ha\_corna} 'has\_horns' and \textit{è\_velenoso} 'has\_poison'}
\label{tab:6}
\begin{tabular}{@{}lll@{}}
\toprule
\textbf{Prototypes} & \textbf{Similarity values} & \textbf{Percentage Similarity} \\
\midrule
\textit{cervo} 'deer' & 0.56 & 6.72 \\
\textit{giraffa} 'giraffe' & 0.07 & 1.38 \\
\textit{mucca} 'cow'& 0.19 & 3.43 \\
\textit{rinoceronte} 'rhinoceros'& 0.45 & 5.1 \\
\textit{toro} 'bull'& 0.37 & 4.48 \\
\hline
\textit{ape} 'bee'& 0.02 & 0.62 \\
\textit{pitone} 'python'& 0.15 & 1.73 \\
\textit{serpente} 'snake'& 0.27 & 3.24 \\
\textit{ragno} 'spider'& 0.05 & 0.91 \\ 
\textit{rana} 'frog'& 0.18 & 1.97 \\
\botrule
\end{tabular}
\end{table}

Table \ref{tab:7} shows the index of Feature association for the word \textit{bue} and the two features. With the system described in this section, we assign a feature to an NPA based on the similarity with prototypes related to that feature and use the relation of the NPA with all the other prototypes. We calculated the score as a difference between the average similarity of related nouns and the average similarity of unrelated nouns.\\ 

\begin{table}[h]
\caption{Index of feature association of ''bue'', \textit{ox}, and two features}
\label{tab:7}
\begin{tabular}{@{}llllll@{}}
\toprule
\textbf{Feature} & \textbf{Prototypes} & \textbf{$S_{rel}$} & $S_{unrel}$ & $C_t$ & $F_t$\\
\midrule
\textbf{has\_poison} & ape, serpente, pitone, ragno, rana & 1.62 & 2.01 & 0 & -0.39\\
\textbf{has\_horns} & cervo, toro, mucca, rinoceronte, giraffa & 4.22 & 1.72 & 0.68 & 3.19\\
\botrule
\end{tabular}
\end{table}

This technique is similar to the strategy adopted by \cite{pado2007dependency} and \cite{mcdonald2004distributional} in their Single-Word Priming experiment in which the model must select the correct prime word from a list of words. In addition, we propose another element, Centrality, that helps us to emphasise the importance of highly related words. The Centrality values consider the number of words that have a similarity with the target word significantly higher than the average (about 2) in relation to the number of words associated with the specific feature.

The last step regards the definition of a threshold, the parameter \textit{PK}, over which a feature is assigned to the description of the NPA noun.

\subsubsection{Experimental Results}
\label{experiment}
To test the accuracy of the method, we planned an experimental step characterised by three phases:

\begin{itemize}
    \item Identify a set of animal nouns included in the base-map of our matrices with a medium frequency value in the training corpus. For this step, we started from the list of animals used to generate the animal network (see section \ref{experiment}), excluding animals with too low frequencies. Then, we try to include animals with disparate characteristics to include as many features as possible in our experiment.
    \item Test our method with different thresholds to obtain the best description and maximise the accuracy.
    \item Test the same model with a single generic matrix and compare the results.
\end{itemize}

The first step brings us to define a list of 67 NPA nouns which belong to the animal classes of mammals, reptiles, amphibious, fishes, birds and molluscs. We select those animals by considering the frequency of their nouns in the corpus. We used the frequency list of the paisà corpus and selected the nouns with more than 100 occurrences. We excluded nouns of sub-species (such as \textit{bonobo}, \textit{dobermann} or \textit{labrador}), but also nouns of animals such as \textit{scricciolo} ’wren’ or \textit{cardellino} ’goldfinch’. The final list of 67 animals is the following:

\begin{multicols}{3}
    \begin{itemize}
    \scriptsize
\item[]	acciuga, \textit{anchovy}
\item[]	agnello, \textit{lamb}
\item[]	alligatore, \textit{alligator}
\item[]	anatra, \textit{duck}
\item[]	antilope, \textit{antelope}
\item[]	aragosta, \textit{lobster}
\item[]	babbuino, \textit{baboon}
\item[]	biscia, \textit{grass snake}
\item[]	bisonte, \textit{bison}
\item[]	bue, \textit{ox}
\item[]	bufalo, \textit{buffalo}
\item[]	calabrone, \textit{hornet}
\item[]	cane, \textit{dog}
\item[]	capra, \textit{goat}
\item[]	cavalletta, \textit{grasshopper}
\item[]	cicogna, \textit{stork}
\item[]	cinghiale, \textit{boar}
\item[]	cobra, \textit{cobra}
\item[]	corvo, \textit{crow}
\item[]	dinosauro, \textit{dinosaur}
\item[]	dromedario, \textit{dromedary}
\item[]	falco, \textit{falcon}
\item[]	foca, \textit{seal}
\item[]	gallo, \textit{rooster}
\item[]	gambero, \textit{shrimp}
\item[]	gazza, \textit{magpie}
\item[]	gazzella, \textit{gazelle}
\item[]	geco, \textit{gecko}
\item[]	gorilla, \textit{gorilla}
\item[]	gru, \textit{crane}
\item[]	iena, \textit{hyena}
\item[]	leopardo, \textit{leopard}
\item[]	lepre, \textit{hare}
\item[]	libellula, \textit{dragonfly}
\item[]	lucertola, \textit{lizard}
\item[]	maiale, \textit{pig}
\item[]	mosca, \textit{fly}
\item[]	mulo, \textit{mule}
\item[]	oca, \textit{goose}
\item[]	orca, \textit{orca}
\item[]	papero, \textit{gosling}
\item[]	pavone, \textit{peacock}
\item[]	pecora, \textit{sheep}
\item[]	pettirosso, \textit{robin}
\item[]	piccione, \textit{pigeon}
\item[]	pony, \textit{pony}
\item[]	puma, \textit{puma}
\item[]	ratto, \textit{rat}
\item[]	rondine, \textit{swallow}
\item[]	rospo, \textit{toad}
\item[]	sanguisuga, \textit{leech}
\item[]	sardina, \textit{sardine}
\item[]	scarafaggio, \textit{beetle}
\item[]	scoiattolo, \textit{squirrel}
\item[]	seppia, \textit{cuttlefish}
\item[]	stambecco, \textit{ibex}
\item[]	struzzo, \textit{ostrich}
\item[]	tacchino, \textit{turkey}
\item[]	tarantola, \textit{tarantula}
\item[]	tonno, \textit{tuna}
\item[]	tordo, \textit{thrush}
\item[]	trota, \textit{trout}
\item[]	verme, \textit{worm}
\item[]	vespa, \textit{wasp}
\item[]	vipera, \textit{viper}
\item[]	zanzara, \textit{mosquito}
\item[]	zebra, \textit{zebra}
    \end{itemize}
    \end{multicols}

We applied our methodology by testing different values of \textit{CK} and \textit{PK} to find the configuration that obtains the best results. The model obtains the best accuracy (F1) score of 0.678 with  $PK=0.71$ and $CK=3.9$.

\begin{figure}[h]
\centering
\includegraphics[width=\textwidth]{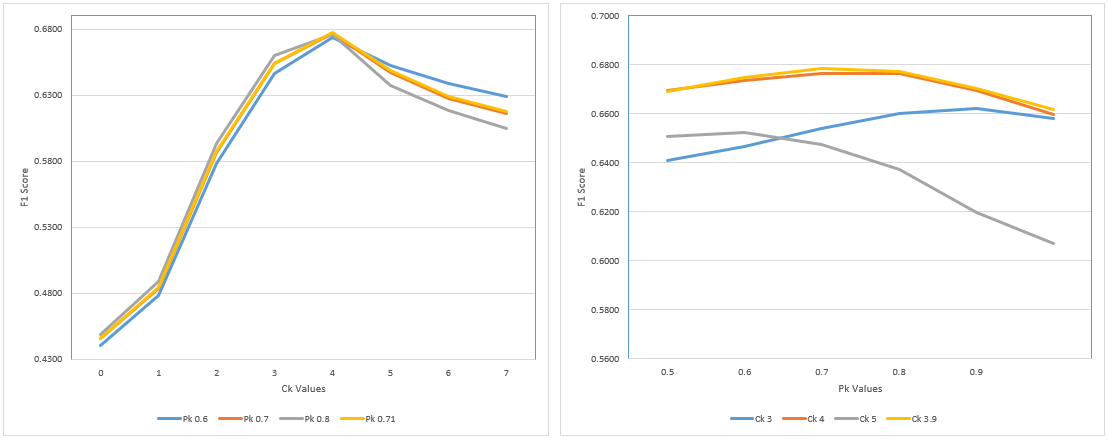}
\caption{Variation of results with different values of parameter CK and PK: on the left, we can observe the curves of results of PK values 0.6, 0.7, 0.71 and 0.8 related to changes in CK; on the right, we can see the curves of F1 scores of CK values 3, 3.9, 4 and 5 in relation with changes in PK}
\label{fig3}
\end{figure}

Figure \ref{fig3} compares F1 scores in relation to changes in values of PK and CK parameters. The curve of results of different PK parameters looks very similar and reaches their tops with a CK included between 3 and 4. The figure on the right shows the curve of results for three different CK values related to changes in PK. In this case, we can observe a similarity among the shapes of the three curves, but with different amplitudes and scores. We observe the main difference in the Ck 3 curve, which reaches the highest point with Pk 0.9.

The optimized model reaches the best score on a single animal with the noun \textit{wasp} (0.85 Fscore) with 17 True Positives, 2 False Positives and 4 False Negatives.

Regarding the feature classes, the best are "Category" and "Number of Legs", with F1 scores of 0.85 and 0.84, respectively. The worst feature class is "Dimension" (0.49).

To establish a benchmark for our methodology, we applied the same formulas outlined in section \ref{method}, employing a single matrix to encompass all identified feature families. We conducted tests using both Word2Vec and BERT, initially seeking to identify the optimized model for each. The optimal outcome for Word2Vec was achieved with a Pk of 0 and a CK of 2.9, whereas the best-performing BERT model was identified with a Pk of 1.4 and a CK of 2.

\begin{table}[]
    \centering
        \caption{Comparison among the recall, precision and F1 Score of the three optimized models}
    \begin{tabular}{l|l|l|l|l|l}
         \textbf{Model} & \textbf{PK} & \textbf{CK} & \textbf{Recall} & \textbf{Precision} & \textbf{F1 Score} \\
         \hline
         SD-W2 & 0.71 & 3.9 & 0.62 & 0.74 & 0.68 \\
         W2V & 0 & 2.9 & 0.37 & 0.74 & 0.50 \\
         BERT & 1.4 & 2 & 0.43 & 0.71 & 0.53 \\
         \hline
         
    \end{tabular}
    \label{featscore}
\end{table}

This difference between the three models appears more clearly by analysing the precision and recall scores. Our optimized model has a precision of 0.75 and a recall of 0.62. In general, those two values have an opposite trend: precision shows shallow values with small CK values, increases significantly with medium values and is stable with higher values. Contrariwise, the Recall curve is very high with lower values of CK and gradually decreases with values higher than 3. The best values of the F1 score are located at the point at which the two curves cross. Nevertheless, with higher values of CK, the F1 score decreased slowly and became stable because the more the CK value increases, the less it produces a variation in feature selection.\\
Unlike the proposed model, W2V and BERT maximize their results with a very low recall (0.37 and 0.43 respectively) but a high score of precision (0.74 and 0.71). When searching for the best Recall score, the precision decreases significantly, reaching a maximum score of 0.25 and 0.34. In conclusion, despite having a similar precision score, the proposed model surpassed the other tested models due to its higher recall score. Essentially, our model is capable of identifying a greater number of features with a high level of precision.

\section{Conclusion}
This paper presented a semantic resource studied as a possible solution to the lack of interpretability of distributional semantics matrices.

Generally, this problem is faced by establishing a connection between a manually built semantic resource and a DS matrix. Our idea is to create a set of Distributional Semantics matrices in which the co-occurrence of each word is calculated in the base of a collection of controlled dictionaries. Those dictionaries include a list of nouns and verbs. Noun dictionaries reflect a semantic classification of words based on the semantic traits of concrete nouns identified by \cite{chomsky1965aspects}. Regarding verbs, we collected three classes of semantic predicates identified as Transfer verbs, Psych Verbs and Verbs involving Bodies. We illustrated the steps involved in building our resource and presented DoMa, a User Interface that helps users extract similarity values from groups of words. Matrices have been calculated by using the SD-W2, a syntactic DS model.

In the second part of the paper, we try to demonstrate the effectiveness of our resource with two experiments. In the first experiment, we used different domain matrices to extract vectors of animal nouns. Combining the vectors of different matrices, each representing a specific semantic trait involved in the animal description, in a tensor, we calculated the similarity among tensors of different animals and created a network. Then, we generated animal classes through a sub-network classification algorithm. The classes generated reflect a common-sense knowledge about animals, rather than a network created by merely calculating similarities with a classic Distributional Semantics (DS) model. In fact, our model not only outperforms W2V but also a powerful Large Language Model (LLM) such as GPT-4.

The second experiment regards automatic semantic feature generations. We rely on prototype theory to select features by extracting similarities from non-prototypical animals and prototypes. We used two external resources to build our experiment: DUAL-PECCS to establish a list of prototypical animals and the McRae feature norms dataset to create an initial set of features. We collected 47 prototypical animals described by over 80 features, subdivided into 12 feature families, each connected with a specific domain matrix. We proposed a formula that combines similarities between non-prototypical and prototype nouns to select the correct features that describe the non-prototypical animal. We tested our model on 67 animals, obtaining a best F1 score of 0.68, which overcame the same model's precision using a generic word2vec matrix (0.50) and BERT (0.53).

To verify the applicability of our methodology across different noun classes, we replicated both experiments in the domain of vehicles, a domain completely distinct from animals, which is well understood by non-experts and contains well-defined sub-classes.

\begin{figure}[htb]
    \centering
    \includegraphics[width = \linewidth]{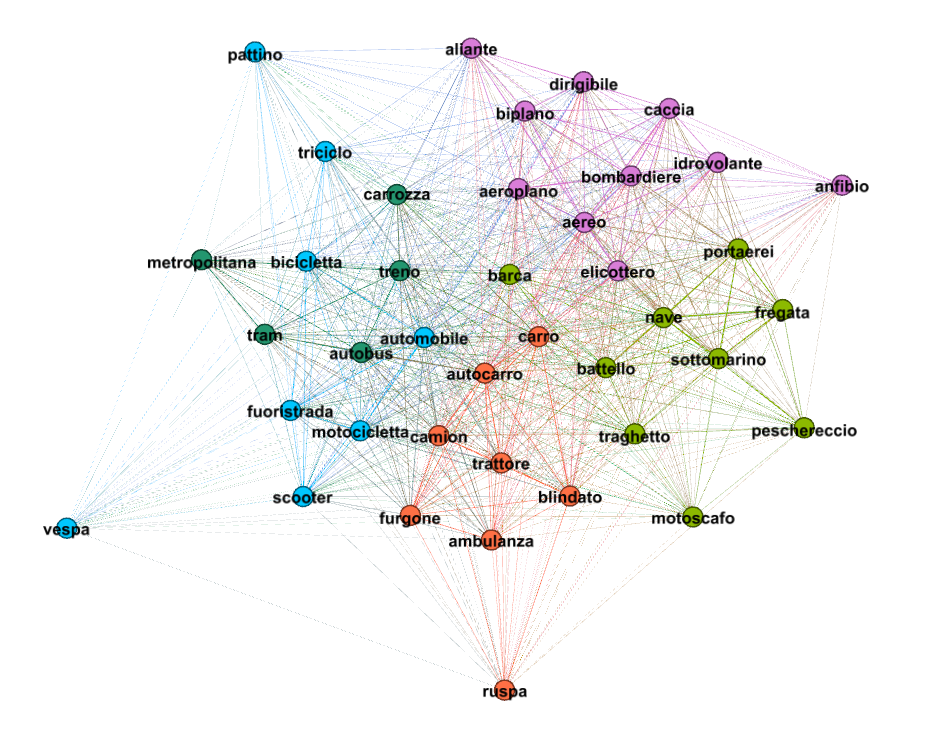}
    \caption{Classification of Vehicles}
    \label{vehi}
\end{figure}

For the initial experiment, we compiled a list of 40 vehicles encompassing a variety of types and functions. We identified seven matrices that could encapsulate the diverse features pertinent to the vehicle class: Vehicles, Human, Locations, Devices, Tools, Materials, and Locative. The classification was carried out as detailed in section \ref{classificationTask}. Figure \ref{vehi} displays the chosen vehicle nouns, the generated graph, and the resulting classification.

Given the limited number of nouns, it is immediately evident that the model provides an effective classification: the five classes can be easily associated with specific categories of vehicles. The fuchsia class encompasses aerial vehicles, albeit with \textit{anfibio} ('amphibious vehicle') incorrectly classified. The orange class comprises work vehicles, while the blue class includes personal vehicles. The dark green class is designated for public transport vehicles, and the light green class exclusively contains water vehicles.

In a quick verification test, we presented the same list of vehicles to ChatGPT-4, which yielded a less accurate classification. The errors made by GPT involved two out of three classes and included the misclassification of \textit{fregata} 'frigate', \textit{portaerei} 'aircraft carrier', \textit{sottomarino} 'submarine', and \textit{metropolitana} 'subway train' as aerial vehicles. Additionally, \textit{anfibi} 'amphibious vehicles' and \textit{dirigibile} 'airship' were incorrectly placed in the class of terrestrial vehicles.

In the second experiment, focused on feature extraction, we manually selected a list of 13 prototypes from the same list of vehicles. Subsequently, we extracted the features for the remaining 27 nouns. The features of vehicles were organized into 5 main classes—Parts, Vector of Motion, Usage, Users, and Manner of Motion—comprising a total of 30 features.

The methodology we proposed achieved a maximum accuracy of 0.672, with a precision of 0.83 and a recall of 0.57.

The experiments validated our proposed resource's ability to bridge the gap between the continuous representation of Distributional Semantics models and the discrete representation of classic semantics. Our approach suggests that by incorporating taxonomic knowledge from domain-specific electronic dictionaries into these models, we can partly address the issue of their lack of connection with the external world—a problem commonly faced by both Distributional Semantic Models and Large Language Models. Looking forward, we aim to enhance the resource by including domain matrices related to semantic predicates not yet utilized, as well as to a sub-classification of abstract nouns.

\bibliography{Semantics}

\end{document}